\documentclass{article}

\usepackage[nonatbib, final]{neurips_2022}

\usepackage[utf8]{inputenc} 
\usepackage[T1]{fontenc}    
\usepackage{hyperref}       
\usepackage{url}            
\usepackage{booktabs}       
\usepackage{amsfonts}       
\usepackage{nicefrac}       
\usepackage{microtype}      
\usepackage[dvipsnames]{xcolor}         
\usepackage{amsmath}
\usepackage{graphicx}
\usepackage{graphics}
\usepackage{subcaption}
\usepackage{wrapfig}
\usepackage{verbatim}
\usepackage[numbers,sort&compress]{natbib}
\usepackage{multirow}
\usepackage{color}
\usepackage{hyperref}
\hypersetup{colorlinks,linkcolor={blue},citecolor={blue},urlcolor={red}}

\usepackage{enumitem}

\usepackage[capitalise]{cleveref}
\usepackage{fancyvrb}

\usepackage[colorinlistoftodos]{todonotes}

\usepackage{tikz}
\usetikzlibrary{fit,positioning}

\usepackage{dsfont}

\usepackage{graphicx} %
\usepackage{amssymb}
\usepackage{times}
\usepackage{epsfig}
\usepackage{graphicx}
\usepackage{color}
\usepackage{bbm}
\usepackage{multicol}
\usepackage{natbib}
\usepackage{wrapfig}

\providecommand{\hide}[1]{}

\usepackage{rotating}

\usepackage{amsthm}
\usepackage{amsopn,amssymb,thmtools,thm-restate}

\usepackage{bm} %
\usepackage{algorithm}%
\usepackage{algpseudocode}%
\newcommand{\vct}[1]{\boldsymbol{#1}} %

\newcommand{\field}[1]{\mathbb{#1}}
\newcommand{\R}{\field{R}} %
\newcommand{\T}{^{\top}} %

\newcommand{\ProbOpr}[1]{\mathbb{#1}}

\newcommand{\expect}[2]{%
\ifthenelse{\equal{#2}{}}{\ProbOpr{E}_{#1}}
{\ifthenelse{\equal{#1}{}}{\ProbOpr{E}\left[#2\right]}{\ProbOpr{E}_{#1}\left[#2\right]}}} %
\newcommand{\var}[2]{%
\ifthenelse{\equal{#2}{}}{\ProbOpr{VAR}_{#1}}
{\ifthenelse{\equal{#1}{}}{\ProbOpr{VAR}\left[#2\right]}{\ProbOpr{VAR}_{#1}\left[#2\right]}}} %

\newcommand{\vx}{{\vct{x}}}

\newcommand{\vh}{\vct{h}}

\newcount\Comments  %
\newcommand{\kibitz}[2]{\ifnum\Comments=1\textcolor{#1}{#2}\fi}
\usepackage{color}
\definecolor{red}{rgb}{1,0,0}

\usepackage{xspace} 
\newcommand{\model}{{\textsc{OptFormer}}\xspace} 

\newcommand{\modelh}{\textsc{OptFormer-H}\xspace}

\newcount\Edits  %
\newcommand{\edit}[1]{\ifnum\Edits=1\textcolor{blue}{#1}\else{#1}\fi}

\newcommand{\vizierdata}{{RealWorldData}\xspace}
\newcommand{\bbobdata}{{BBOB}\xspace}
\newcommand{\hpobdata}{{HPO-B}\xspace}
\newcommand{\ignore}[1]{}

\title{Towards Learning Universal Hyperparameter Optimizers with Transformers}

\newcommand\blfootnote[1]{%
  \begingroup
  \renewcommand\thefootnote{}\footnote{#1}%
  \addtocounter{footnote}{-1}%
  \endgroup
}

\author{Yutian Chen$^{1}$, Xingyou Song$^{2}$, Chansoo Lee$^{2}$, Zi Wang$^{2}$, Qiuyi Zhang$^{2}$, \\
\textbf{David Dohan$^{2}$, Kazuya Kawakami$^{1}$, Greg Kochanski$^{2}$,} \\
\textbf{Arnaud Doucet$^{1}$, Marc'aurelio Ranzato$^{1}$, Sagi Perel$^{2}$, Nando de Freitas$^{1}$} \\
$^1$Deepmind, $^2$Google Research, Brain Team  
}

\begin{document}
\maketitle
\vspace{-0.65cm}
\begin{abstract}
Meta-learning hyperparameter optimization (HPO) algorithms from prior experiments is a promising approach to improve optimization efficiency over objective functions from a similar distribution. However, existing methods are restricted to learning from experiments sharing the same set of hyperparameters. In this paper, we introduce the \model, the first text-based Transformer HPO framework that provides a universal end-to-end interface for jointly learning policy and function prediction when trained on vast tuning data from the wild, such as Google's Vizier database, one of the world's largest HPO datasets. Our extensive experiments demonstrate that the \model can simultaneously imitate at least 7 different HPO algorithms, which can be further improved via its function uncertainty estimates. Compared to a Gaussian Process, the \model also learns a robust prior distribution for hyperparameter response functions, and can thereby provide more accurate and better calibrated predictions. This work paves the path to future extensions for training a Transformer-based model as a general HPO optimizer.
\end{abstract}

\vspace{-0.85cm}

\blfootnote{Code: \url{https://github.com/google-research/optformer}. Google AI Blog: \url{https://ai.googleblog.com/2022/08/optformer-towards-universal.html}.}
\section{Introduction}
\label{sec:introduction}

The emergence of public machine learning data platforms such as OpenML \citep{vanschoren2014openml} and hyperparameter optimization (HPO) services such as Google Vizier \cite{vizier}, Amazon SageMaker \cite{perrone2020amazon} and Microsoft Azure \cite{mukunthu2019practical} have made large-scale datasets containing hyperparameter evaluations accessible. For our use-case in this paper, Google Vizier is the de-facto HPO service across Google, having optimized some of Google's largest products and research efforts, and contains a collection of valuable tuning data within the last 5 years. 
While there is growing interest in leveraging such data to meta-learn hyperparameter optimization algorithms \citep{hpo-b, dnn_bench, wistuba2020few, feurer2018scalable}, dealing with large datasets consisting of experimental trials in the wild can be challenging, due to large variations in HPO problems and their associated text metadata (e.g. shown later in Table \ref{tab:orig_study}).

Thus, most meta and transfer-learning HPO methods \citep{swersky2013multi, yogatama2014efficient, poloczek2017multi, perrone2018scalable, feurer2018scalable, wistuba2020few, rothfuss2021pacoh, krause2011contextual, bardenet2013collaborative, poloczek2016warm} consider a restrictive setting where all tasks must share the same set of hyperparameters so that the input data can be represented as fixed-sized vectors. Consequently, such methods only exploit a small portion of the available data to learn priors. This drawback is more severe for large datasets which contain significant amounts of useful information.

To overcome these limitations, we introduce the \model, a general hyperparameter optimization framework based on Transformers \citep{transformer}. Transformers have demonstrated excellent performance in many data tasks, ranging from natural language \citep{bert}, images \citep{vision_Transformer, image_gpt}, biological data \citep{bio_transformer, bio_transformer2}, code \citep{alphacode, codex}, and control \citep{decision_transformer,reed2022gato}. Here, we investigate how to use a Transformer as a universal interface for modelling experimental data and learn HPO algorithms, as given a sufficient amount of data, a Transformer can potentially learn a more complex prior distribution than standard Bayesian Optimization (BO) with Gaussian Processes (GPs), especially as the Transformer possesses certain computational advantages over GPs for large datasets.

\begin{figure}[t]
\vspace{-0.4cm}
    \centering
    \includegraphics[width=.8\textwidth]{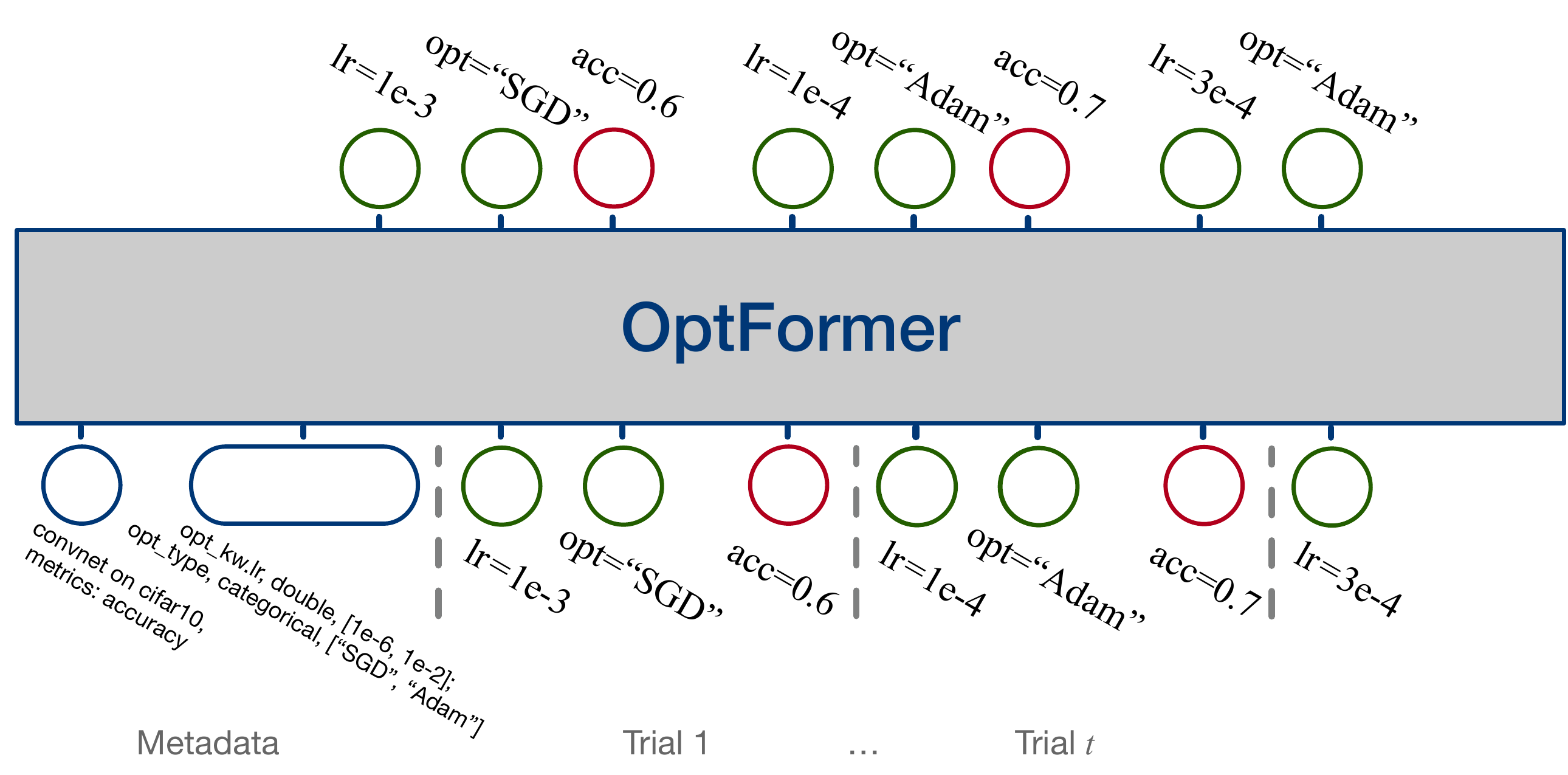}
    \caption{Illustration of the \model model over a hyperparameter optimization trajectory. It is trained to predict both hyperparameter suggestions (in green) and response function values (in red).}
    \label{fig:model_graph}
\end{figure}

We introduce a serialization scheme to convert a combination of any metadata and an optimization trajectory into text, represented as a sequence of tokens, and formulate the HPO task as a sequence modeling problem. We adopt a supervised learning approach, by learning to predict parameters and hyperparameter response functions from offline tuning data (See \cref{fig:model_graph}). In order to further improve optimization performance, we augment the model by utilizing its own function prediction during inference (\cref{sec:inference}). Extensive experiments on both public and private datasets demonstrate the \model's competitive tuning and generalization abilities.

In summary, our contributions are as follows:
\begin{itemize}[leftmargin=*]
\item We formulate, to the best of our knowledge, the first meta-learning HPO framework to learn both \textbf{policy} and \textbf{function priors} from data across different search spaces.
\item The \model is capable of learning the behaviors of 7 diverse blackbox optimization algorithms relying on a broad class of methods (non-adaptive, evolutionary, and Bayesian).
\item Furthermore, the \model learns the prior over objective functions and provides both accurate and well calibrated predictions, in many cases significantly surpassing GPs in log-predictive likelihood and expected calibration error (ECE)~\citep{naeini2015obtaining}.
\item Lastly, \model policies augmented with model-based optimization, such as the use of Expected Improvement acquisition functions, are competitive HPO algorithms. To the best of our knowledge, this is the first time Transformers are augmented with acquisition functions for online adaptation.
\end{itemize}

\section{Preliminaries}

\subsection{Meta-learning for hyperparameter optimization}
HPO aims to find\ignore{The goal of HPO is to tune} a set of hyperparameters $\vx$ from search space $\mathcal{X}$ to maximize a model performance metric, $y=f(\vx)$, often referred to as a response function. \cref{tab:orig_study} shows an example of HPO experimental data. Following the HPO nomenclature \citep{vizier, raytune}, an experimental study consists of metadata ($m$) and a history of trials ($\vh$). \ignore{an experimental \textit{study} $=(m, h)$, includes both metadata $m$ \ignore{\textit{metadata} $m$ 
(\textcolor{blue}{blue})} and a history of trials $h$.\ignore{\textit{history} $h$ (\textcolor{purple}{purple}) of trials.}} The metadata contains arbitrary unstructured information, including but not limited to descriptions of the problem, optimization algorithm, names, types and value ranges of hyperparameters. The history after $t$ trials, $\vh_t = \left(\vx_{1}, y_{1}, \dots, \vx_{t}, y_{t}\right)$, contains a sequence of trials, each of which consists of a parameter suggestion $\vx$ and function value $y$\ignore{ in a tuple $(\vx, y)$}.

The goal of the meta-learning approach for HPO is to learn the shared knowledge among the objective functions $f$ from a dataset of multiple tuning experiments represented as studies and to obtain an optimal HPO algorithm for new hyperparameter tuning tasks from a similar distribution to those in the dataset.
\ignore{
Given a dataset $\mathcal{D}$ of multiple tuning experiments represented as studies, the goal of the meta-learning approach is to learn the shared knowledge among the objective functions $f$ in the data and obtain an optimal HPO algorithm for new hyperparameter tuning tasks from a similar distribution to those in $\mathcal{D}$.}

An HPO algorithm $\pi$ maps the metadata and history to a distribution over hyperparameter suggestions, i.e. $\pi(\vx_{t+1}|m, \vh_{t})$. Using the terminology of offline RL \citep{levine2020tutorial}, we refer to the algorithm used to generate the trajectories in a dataset as the behavior policy $\pi_b$.

We primarily consider search spaces $\mathcal{X}$ with a fixed number $D$ of hyperparameters per task, and hence $\vx = (x^{(1)},\dots, x^{(D)})$, with each hyperparameter $x^{(d)}$ being of type \texttt{DOUBLE}, \texttt{INTEGER}, \texttt{DISCRETE}, or \texttt{CATEGORICAL} (see \cref{appendix:ss_primitives} for details). More complex search spaces can be supported as discussed in \cref{sec:limitations}.
\ignore{
We primarily consider search spaces $\mathcal{X}$ with a fixed number $D$ of hyperparameters per task, and hence $\vx = (x^{(1)},\dots, x^{(D)})$, with each hyperparameter $x^{(d)}$ being of type \texttt{Double}, \texttt{Integer}, \texttt{Discrete}, or \texttt{Categorical} (see \cref{appendix:ss_primitives} for details). More complex search spaces can be supported as discussed in \cref{sec:limitations}.}

\begin{wraptable}[31]{r}{0.34\textwidth}
\vspace{-36pt}
\caption{Example of a study $(m, \vh)$ with two parameters and two trials. Metadata $m$ appears in \ignore{\textcolor{blue}{blue}}blue and history $\vh$ in \ignore{\textcolor{purple}{purple.}}purple.}
\begin{tabular}{l} \toprule
\textcolor{blue}{\scriptsize \tt \bf  "name": "convnet on cifar10",}\\
\textcolor{blue}{\scriptsize \tt \bf    "metric": "accuracy",}\\
\textcolor{blue}{\scriptsize \tt \bf   "goal": "MAXIMIZE",}\\
\textcolor{blue}{\scriptsize \tt \bf   "algorithm": "random\_search",}\\
\textcolor{blue}{\scriptsize \tt \bf   "parameter": \{}\\
\textcolor{blue}{\scriptsize \tt   "name": "opt\_kw.lr",}\\
\textcolor{blue}{\scriptsize \tt   "type": "DOUBLE",}\\
\textcolor{blue}{\scriptsize \tt   "min\_value": 1e-6,}\\
\textcolor{blue}{\scriptsize \tt   "max\_value": 1e-2,}\\
\textcolor{blue}{\scriptsize \tt   "scale\_type": "LOG"}\\
\textcolor{blue}{\scriptsize \tt  \bf  \}}\\
\textcolor{blue}{\scriptsize \tt \bf   "parameter": \{}\\
\textcolor{blue}{\scriptsize \tt   "name": "opt\_type",}\\
\textcolor{blue}{\scriptsize \tt   "type": "CATEGORICAL",}\\
\textcolor{blue}{\scriptsize \tt   "categories": ["SGD", "Adam"],}\\
\textcolor{blue}{\scriptsize \tt \bf   \}}\\
\textcolor{purple}{\scriptsize \tt \bf  "trial" \{}\\
\textcolor{purple}{\scriptsize \tt   "parameter": \{}\\
\textcolor{purple}{\scriptsize \tt     "opt\_kw.lr": 0.0021237573,}\\
\textcolor{purple}{\scriptsize \tt     "opt\_type": "SGD"}\\
\textcolor{purple}{\scriptsize \tt   \}}\\
\textcolor{purple}{\scriptsize \tt   "metric": \{}\\
\textcolor{purple}{\scriptsize \tt     "accuracy": 0.69482429,}\\
\textcolor{purple}{\scriptsize \tt   \}\bf\}}\\
\textcolor{purple}{\scriptsize \tt \bf  "trial" \{}\\
\textcolor{purple}{\scriptsize \tt   "parameter": \{}\\
\textcolor{purple}{\scriptsize \tt     "opt\_kw.lr": 0.00038292234,}\\
\textcolor{purple}{\scriptsize \tt     "opt\_type": "Adam"}\\
\textcolor{purple}{\scriptsize \tt   \}}\\
\textcolor{purple}{\scriptsize \tt   "metric": \{}\\
\textcolor{purple}{\scriptsize \tt     "accuracy": 0.71642583}\\
\textcolor{purple}{\scriptsize \tt   \}\bf\}}\\
\bottomrule
\end{tabular}
\vspace{8mm}
\label{tab:orig_study}
\end{wraptable}

\subsection{Transformer model}
The Transformer model is an efficient attention-based neural network architecture for sequence modeling \citep{transformer}. We adopt the T5 Transformer encoder-decoder architecture \citep{2020t5}. The encoder and decoder each consist of a stack of multi-head self-attention layers which construct pairwise interactions between positions, followed by position-wise feed-forward networks. The encoder converts a sequence of input token representations $m$, to a sequence of continuous embeddings, which is fed to the decoder to generate a sequence of output tokens $h$ one element at a time (see \cref{fig:model_graph}).

\section{Related work}
\label{sec:related_work} 

There has been a rich set of works in meta-learning and transfer learning by modifying specific core components of the BO pipeline, such as the acquisition function or the GP, in order to tackle BO's myopic behavior, or obtaining more information from similar tasks. For instance, approaches include learning new acquisition functions \citep{metalearn-acquisition}, multi-task BO \citep{swersky2013multi, yogatama2014efficient, poloczek2017multi, perrone2018scalable, feurer2018scalable, wistuba2020few, rothfuss2021pacoh} and BO for transfer learning using contextual GPs \citep{krause2011contextual, bardenet2013collaborative, poloczek2016warm}. \citep{Feurer} also studies the use of meta-BO for hyperparameter tuning tasks in machine learning. However, all of these works consider a fixed search space.

A more radical meta-learning approach to non-differentiable optimization trains recurrent neural networks (RNNs) as neural optimizers from scratch. \citep{chen2017learning} first proposed training an RNN with gradient descent to optimize blackbox functions and hyperparameters while \citep{rl_squared,l2rl} train RNNs using reinforcement learning (RL) to solve RL tasks. Unfortunately, prior works are limited to fixed search spaces and only use online generated data, constraining the training objectives to be cheaply computable.

In this work, we wish to overcome the limitations of previous works by exploiting the Transformer architecture. Numerous works have demonstrated Transformers' strong capabilities in flexible symbolic and numerical manipulation. On the symbolic side, Transformers have been shown able to manipulate symbolic mathematical expressions \citep{symbolic_gpt, symbolic_math_Transformer, polynomial_simplification_Transformer} and generate code \citep{alphacode, codex}. Furthermore, on the numerical side, Transformers have also been shown able to perform linear algebra computations \citep{linear_algebra_Transformers}, Bayesian Inference \citep{gp_transformer}, and offline RL \citep{online_decision_transformer, decision_transformer,reed2022gato}. For AutoML specifically, \citep{privileged_zeroshot_automl} has demonstrated Transformers' and analogous graph neural networks' abilities to use dataset descriptions and metadata to generate classification and data preprocessing pipelines. However, to date, there has been little effort in attacking the full problem of hyperparameter tuning in the blackbox optimization setting. In this paper, the challenging task of learning algorithms from blackbox optimization trajectories can be seen as a significant extension of both symbolic and numerical manipulation. Since the underlying algorithm can be composed of multiple symbolic and mathematical operations with unbounded complexity, the model must infer potentially very complex behavior over long horizons.

\section{Universal interface and model for hyperparameter optimization}

In this section, we provide a universal interface for modeling HPO studies with mixed textual and numerical information as a sequence of discrete tokens. We train our \model as a generative model on a given dataset and explain how to use the \model's parameter and function prediction abilities to implement an HPO policy.

\subsection{Study tokenization}
\label{subsec:preprocessing}
To generalize over HPO problems of different parameter sizes, types, and metadata, we propose to serialize the study as a one-dimensional textual sequence, also advocated in \cite{reed2022gato}. Unfortunately, a naive serialization approach, e.g. via JSON \citep{json}, will produce unnecessarily long sequences.

To improve scalability, we compress the textual representation of metadata $m$ by removing redundant phrases and punctuation (e.g., "parameter", quotes) and encoding keywords (e.g., "name", "algorithm") and enumerating types (e.g. "DOUBLE") into single tokens.

For the historical sequence $\vh$, we convert every \texttt{DOUBLE} and \texttt{INTEGER} parameter along with every function value into a single token, by normalizing and discretizing them into integers, with an quantization level of $Q=1000$; e.g.
\begin{equation}
\bar{x} = \mathrm{int} [x_{\mathrm{norm}} \cdot Q], \text{ where }
x_{\mathrm{norm}} = (x - x_\mathrm{min})/(x_\mathrm{max} - x_\mathrm{min}).    
\end{equation}
The range of $x$ is defined by the search space and the range of $y$ is obtained from observed values in $\vh$. For other types, we use the index in their value set.

The shortened text string is then converted to a sequence of tokens via the SentencePiece tokenizer \citep{kudo2018sentencepiece} (see \cref{tab:preprocessed_study} for an example). Every trial is represented by text, which is represented as a sequence of normalized and quantized tokens, $\left[\bar{x}_t^{(1)}, \dots, \bar{x}_t^{(D)}, \star, \bar{y}_t, \text{"|"}\right]$, where the token $\star$ separates parameter and function values and "|" separates trials. See \cref{sec:details_preprocessing} for further details on tokenization.

\begin{table}[t]
\caption{Serialized study after preprocessing and tokenization. Metadata $m$ appears in blue\ignore{\textcolor{blue}{blue}}, normalized and quantized values of $\vx_t$ in green\ignore{\textcolor{ForestGreen}{green}}, and $y_t$ in red\ignore{\textcolor{red}{red}}.} 
\label{tab:preprocessed_study}
\begin{tabular}{p{1.1in}|p{4in}} \toprule
After preprocessing &
\begin{minipage}{3in}
{\scriptsize
\begin{Verbatim}[commandchars=\\\{\}]
\textcolor{blue}{ \bf  <name>:"convnet on cifar10",<metric>:"accuracy",<goal>:<MAXIMIZE>,}
\textcolor{blue}{ \bf  <algorithm>:"random_search"}
\textcolor{blue}{ \bf  &<name>:"opt_kw.lr",<type>:<DOUBLE>,<min_value>:1e-6,<max_value>:1e-2,}
\textcolor{blue}{ \bf  <scale_type>:<LOG>}
\textcolor{blue}{ \bf  &<name>:"opt_type",<type>:<CATEGORICAL>,<categories>:["SGD", "Adam"]}
\textcolor{ForestGreen}{ \bf <831><0>}*\textcolor{red}{\bf <0>}|\textcolor{ForestGreen}{\bf <645><1>}*\textcolor{red}{\bf <999>}
\end{Verbatim}
}
\end{minipage}
\\
\midrule
 \multirow{1}{*}[0.34em]{\hspace{0.12cm}\shortstack{Subwords \\ after tokenization}} &
\begin{minipage}{3in}
{\scriptsize
\begin{Verbatim}[commandchars=\\\{\}]
\textcolor{blue}{ \bf  name : " con v net on ci far 10 ", metric : " acc u racy ",}
\textcolor{blue}{ \bf  goal : MAXIMIZE , algorithm : " random _ search "}
\textcolor{blue}{ \bf  & name : " op t _ kw . lr ", type : DOUBLE , min_value : 1 e -6 , }
\textcolor{blue}{ \bf  max_value : 1 e -2 , scale_type : LOG}
\textcolor{blue}{ \bf  & name : " op t _ type ", type : CATEGORICAL , }
\textcolor{blue}{ \bf  categories : [ " SG D ", " A dam " ]}
\textcolor{ForestGreen}{ \bf 831 0} * \textcolor{red}{\bf 0} | \textcolor{ForestGreen}{\bf 645 1} * \textcolor{red}{\bf 999}
\end{Verbatim}
}
\end{minipage}
\\
\bottomrule
\end{tabular}
\end{table}

\subsection{Model and training loss}
\label{subsec:model}
After tokenization, the converted historical sequence is as follows:
\begin{equation}
\bar{\vh}_t = \left[\bar{x}_1^{(1)}, \bar{x}_1^{(2)}, \dots, \bar{x}_1^{(D)}, \star, \bar{y}_1, \text{"|"}, \dots,
          \bar{x}_t^{(1)}, \bar{x}_t^{(2)}, \dots, \bar{x}_t^{(D)}, \star, \bar{y}_t\right].
\label{eq:barh}          
\end{equation}
We can now apply a Transformer model to learn the conditional distribution of tokens in $\bar{h}$ using the chain rule, given the metadata $\bar{m}$, as depicted in \cref{fig:model_graph}. The joint distribution is presented in \cref{appendix:conditional_probabilities}.

Given a dataset $\mathcal{D}$ of hyperparameter optimization studies, we train the \model by maximizing the weighted log-likelihood for each study $(m, \vh) \sim \mathcal{D}$:
\begin{equation}
 \textstyle   \mathcal{L}(\theta; m, \vh)
    = \sum_n w_n \log P_{\theta}(\bar{h}^{(n)}|\bar{m}, \bar{\vh}^{(1:n-1)}), 
\end{equation}
with $w_n = 0$ if $\bar{h}^{(n)} \in \{\star, \text{"|"}\}$ and $w_n = 1$ otherwise.
That is, we mask out the separator tokens ($\star$, "|") and predict parameter $\bar{\vx}$ and function tokens $\bar{y}$ only. Note that $\bar{h}^{(n)}$ denotes the $n$-th token, that is the $n$-th element of the list in Equation~(\ref{eq:barh}), and $\bar{\vh}^{(1:n-1)}$ denotes all tokens up to the $(n-1)$-th token. Further details and data augmentations are provided in \cref{sec:training_details}.

\subsection{Inference and decoding}
\label{sec:inference}

\paragraph{Parameter prediction:} To decode the predicted parameter token $\bar{x}_t^{(d)}$ 
back to its original parameter range, we truncate the output distribution to the vocabulary range corresponding to valid parameter values $[0, Q)$ and reverse our tokenization procedure in \cref{subsec:preprocessing}. For a \texttt{DOUBLE} or \texttt{INTEGER} parameter $x$, we use a piecewise constant distribution:
\begin{equation}
    p_\theta(x | \dots) = Q \cdot P_\theta(\bar{x} | \dots) / (x_\mathrm{max} - x_\mathrm{min}) ,
    \text{ if } x \in [x_\mathrm{min}, x_\mathrm{max}], \text{ otherwise } 0\,.
    \label{eq:decoding}
\end{equation}
For all other parameter types, $\bar{x}$ corresponds to the index of the set of feasible values. Putting these together, we may now sample parameter $\vx_{t}$ from the model's prior distribution and thus define an HPO policy:
\begin{equation}
    \pi_{\text{prior}}(\vx_t|m, \vh_{t-1}) = \prod_{d=1}^D p_\theta(x_t^{(d)}|m, \vh_{t-1}, \vx_t^{(1:d-1)}).
\end{equation}
As we use a supervised learning loss, we expect $\pi_{\text{prior}}$ to approximate the behavior policy $\pi_b$. 

Note that traditional BO algorithms require running Bayesian inference and then conducting a global search in the hyperparameter space with an acquisition function. Thus the runtime complexity of making one hyperparameter suggestion is cubic in $t$ for a typical GP-based BO method that performs ARD each iteration \citep{garnett_bayesoptbook_2022}. In contrast, generating one suggestion by the \model consists of decoding $D$ parameter tokens with an input sequence of $(D+3)t$ tokens, which are then parsed into the $D$ parameter values, producing a runtime of $\mathcal{O}(D^2 t)$ linear in $t$, with proper caching.

\paragraph{Function prediction:} 
To decode the real-valued function $y_t$ from the discrete distribution $P_\theta(\bar{y}_t|\bar{m}, \bar{\vh}_{t-1}, \bar{\vx}_t)$, we construct the same piecewise constant distribution as in \cref{eq:decoding} with the range $[y_\mathrm{min}, y_\mathrm{max}]$ used in tokenization. Note that the limited support of $y$ will not be a concern for HPO when either the range is known or we set the range large enough compared to observed values. For more general use as a few-shot function prediction model, one could consider adopting the Riemann Distribution in \cite{gp_transformer}, which supports an unbounded range.

\paragraph{Augmented HPO policies with function prediction:} At best, the learned policy $\pi_{\text{prior}}$ can only perform as well as the original policy $\pi_{b}$ when using behavioral cloning. However, we can take advantage of the model's simultaneous function prediction ability to improve the policy with model-based planning or offline RL techniques. While a comprehensive study of policy improvements for Transformers is out of the scope of this work, we consider here a simple yet effective policy improvement operator: sampling $M=100$ candidate suggestions from $\pi_{\text{prior}}$ and choosing the suggestion with the highest score defined by an acquisition function $u(\cdot)$ as follows:
\begin{equation}
    \pi_{u}(\vx_t|m, \vh_{t-1}) = \underset{\{\vx^{(i)}\}_{i=1}^M}{\mathrm{argmax}} ~ u(p_\theta(\cdot|m, \vh_{t-1}, \vx^{(i)})), \text{ with } \vx^{(i)} \overset{\textup{i.i.d.}}{\sim} \pi_{\text{prior}}(\vx|m, \vh_{t-1}).
\end{equation}
Common acquisition functions include Expected Improvement (EI), Probability of Improvement (PI), Upper Confidence Bound (UCB), and Thompson Sampling, see for example \cite{shahriari2015taking}.
\edit{
At a high level, this approach to combining imitated policies with function prediction is reminiscent of the idea behind the offline RL approach of BCQ~\citep{fujimoto2019off}.}

\edit{Because we apply a linear mapping from the original $y$ value to the quantized value $\bar{y}$ before discretization, we can simply define the acquisition functions on the discrete distribution $P_\theta(\bar{y}|\bar{m},\bar{\vh}_{t-1},\bar{\vx}_t)$ as follows:
}
\begin{align}
u_{\text{EI}}(\vx|\bar{y}^*) &= \mathbb{E}_{P_\theta(\bar{y}|m, \vh_{t-1}, \vx)}\left[\max(\bar{y} - \bar{y}^*, 0)\right]\,,\\
u_{\text{UCB}}(\vx|\alpha) &= \mathrm{Quantile}(P_\theta(\bar{y}|m, \vh_{t-1}, \vx_t), \alpha)\,,\\
u_{\text{PI}}(\vx|\bar{y}^*) &= \sum_{\bar{y}>\bar{y}^*} P_\theta(\bar{y}|m, \vh_{t-1}, \vx)\,,\\
u_{\text{TS}}(\vx) &= \bar{y} , \text{ with } \bar{y} \sim P_\theta(\bar{y}|m, \vh_{t-1}, \vx_t)\,,
\end{align}
\edit{where $\bar{y}^* = \max_{\tau \leq t-1} \bar{y}_{\tau}$ in EI and PI is the threshold to measure improvement. We define the UCB acquisition function with a quantile parameter $\alpha$. Our TS acquisition is defined as a sampled function value at a given location from the marginal predictive distribution. It is inspired by the traditional Thompson Sampling method \citep{garnett_bayesoptbook_2022} but different in that the correlation between different locations is ignored.}

\section{Data}
\label{sec:dataset}
Training the \model requires HPO studies with optimization trajectories.
The most natural dataset we possess is the entire Google Vizier \cite{vizier} database, one of the world's largest collections of real world hyperparameter tuning studies, which we denote as \textbf{\vizierdata}. There are around 750K studies, each with on average 300 trials, covering a vast class of production and machine learning applications at Google, ranging from vision, speech, NLP and robotics, and representing one of the most representative distributions of HPO tasks for machine learning models in practice. These studies were generated with a mixture of non-adaptive, evolutionary, and BO algorithms. However, as the dataset does not contain sufficient algorithm information, we have to treat the corresponding behavior policy as a randomly mixed algorithm $\pi_{b}$.

In addition, we create two new datasets based on public benchmarks. \textbf{\hpobdata} is the largest public benchmark for HPO containing about 1.9K tuning tasks, most of which use one of 16 shared search spaces. In the continuous evaluation setting, it fits an XGBoost model to the trial data of every tuning task as the objective function.
For further control over specific function dimensions and properties, we use the blackbox optimization benchmark \textbf{\bbobdata} \citep{ElHara2019COCOTL}, consisting of 24 types of synthetic functions with customizable properties (dimension sizes, rotations, shifts, discretizations, noise types) we randomize over. 

For each of the two public benchmarks (\hpobdata and \bbobdata), we apply a fixed set of 7 HPO algorithms to generate a dataset of optimization trajectories. In contrast to \vizierdata, we specify the algorithm name in the metadata $m$ as part of the conditioning input for our model. The controlled algorithms used are: (1) Grid Search, (2) Shuffled Grid Search, (3) Random Search, (4) Regularized Evolution \cite{real2019regularized}, (5) Hill-Climbing, (6) Eagle Strategy \cite{yang2010eagle}, and (7) Vizier's GP-UCB \cite{vizier}. \cref{appendix:algorithms} contains detailed explanations of the algorithms.

\begin{table}[h]
\begin{center}
\caption{Offline training datasets considered in this study. More details are given in \cref{appendix:data} along with examples of studies in \cref{tab:dataset_study_example}.}
\vspace{0.1cm}
\label{table:dataset_differences}
\begin{tabular}{ |c|c|c|c| } 
 \hline
  & ("R") \textbf{\vizierdata} & ("H") \textbf{\hpobdata} & ("B") \textbf{\bbobdata} \\
  \hline \hline
 \#Studies & 750K &  10M & 10M \\
 \hline
 \#Trials / study & 300 (on average) & 120 & 300 \\
 \hline
 Study source & Google's database &  Generated & Generated \\
 \hline
 $\pi_{b}$ & Mixed &  Controlled & Controlled \\
 \hline
 Obj. Functions & HPO tasks & HPO tasks & Synthetic \\
 \hline
Search space & Different per task & 16 shared search spaces & Randomized \\
 \hline
\end{tabular}
\end{center}
\end{table}

\section{Experiments}
\label{sec:experiments}
We train a single Transformer model with 250M parameters on the union of the three datasets described above, \vizierdata, \hpobdata, and \bbobdata (hyperparameter details in \cref{sec:training_details}).

Each dataset contains a corresponding ``test'' set of functions, either using synthetic functions (\bbobdata) or fitting a machine learning model to obtain the objective (\vizierdata, \hpobdata).
We evaluate mainly on the two natural HPO benchmarks, \vizierdata and \hpobdata. \edit{
The train/test subsets of \vizierdata are split temporally to avoid information leak (see \cref{appendix:data} for details).
}

\edit{
To aggregate results across functions with different output scaling,
we normalize all the test functions. This is standard practice in the literature \cite{hpo-b,turner2021bayesian_bbochallenge,vizier,cowen2022hebo,metz2020taskset,evchenko2021frugal}.} We define our performance metric at trial $t$ as the best-so-far normalized function value $\max_{i\in\{1:t\}} (y_{i} - y_\mathrm{rand})/(y_\mathrm{max} - y_\mathrm{rand})$, where $y_\mathrm{rand}$ is the median of function values randomly sampled in the search space to be robust to outliers, and $y_\mathrm{max}$ is the maximum, if known, or best value found by any algorithm. For the HPO-B benchmark, we use the recommended bounds provided in \citep{hpo-b}. We also consider other metrics when comparing different algorithms in \cref{sec:more_exp_compare_policy}, including the performance profile and average ranking. We find our results are consistent over different metrics.

Because the \model is trained to predict the conditional distributions of parameter and function values, we would like to answer the following questions when evaluating on unseen test problems:

\newpage 
\begin{enumerate}[leftmargin=*]
    \item Can the \model learn to imitate multiple HPO algorithms with one model? (\cref{sec:imitate})
    \item Can the \model learn a good prior over hyperparameter response functions? (\cref{sec:fun_prior})
    \item Is the \model a competitive approach for HPO? (\cref{sec:compare_policy})
\end{enumerate}

\subsection{Imitating HPO policies}
\label{sec:imitate}
\begin{figure}[t]
    \centering
    \begin{subfigure}[b]{0.32\textwidth}
        \centering
        \includegraphics[width=\textwidth]{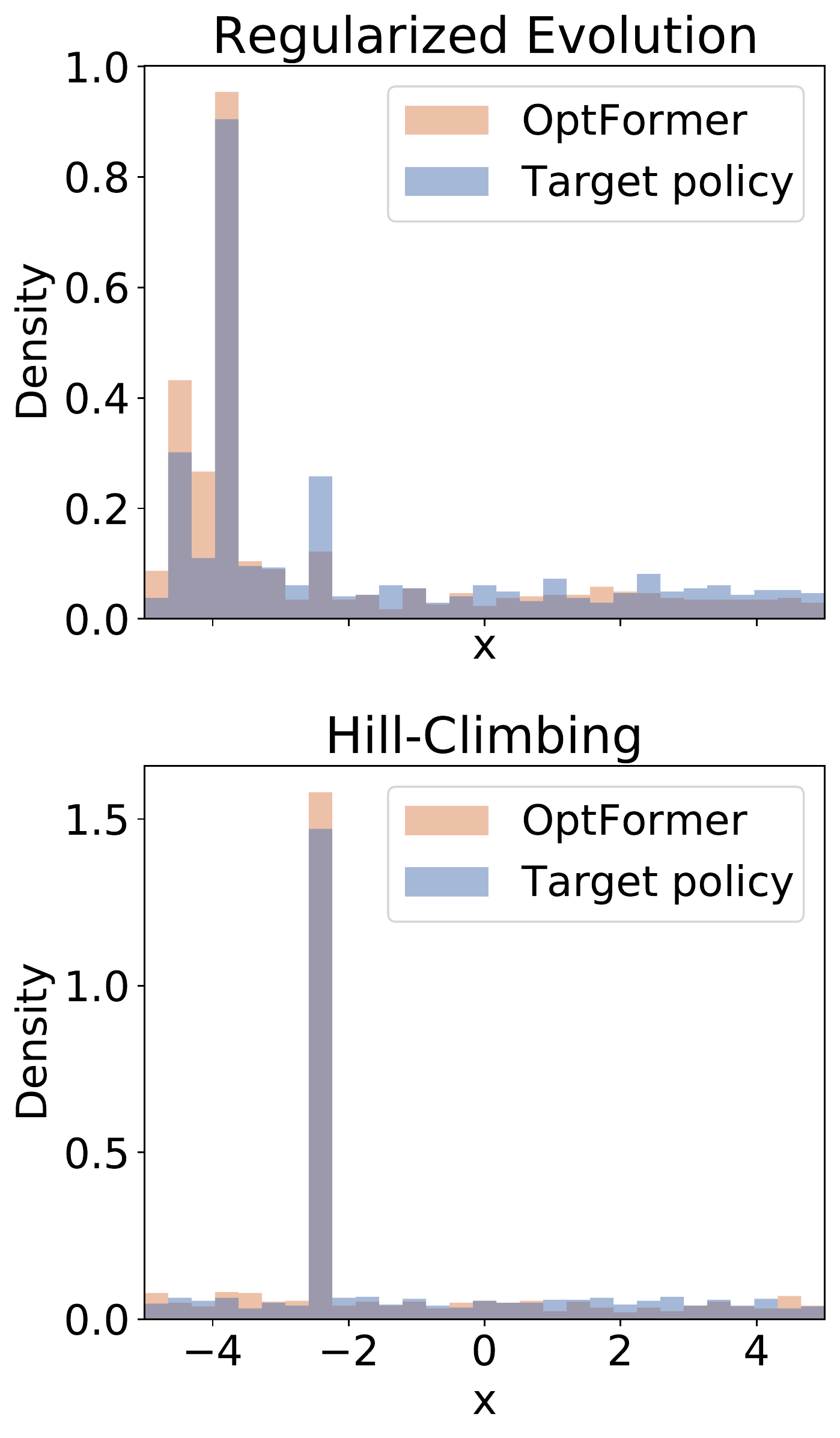}
        \caption{Policy distribution}
        \label{fig:bbob_x_pred_at_trial_40_2alg}
    \end{subfigure}
    \hfill
    \begin{subfigure}[b]{0.32\textwidth}
        \centering
        \includegraphics[width=\textwidth]{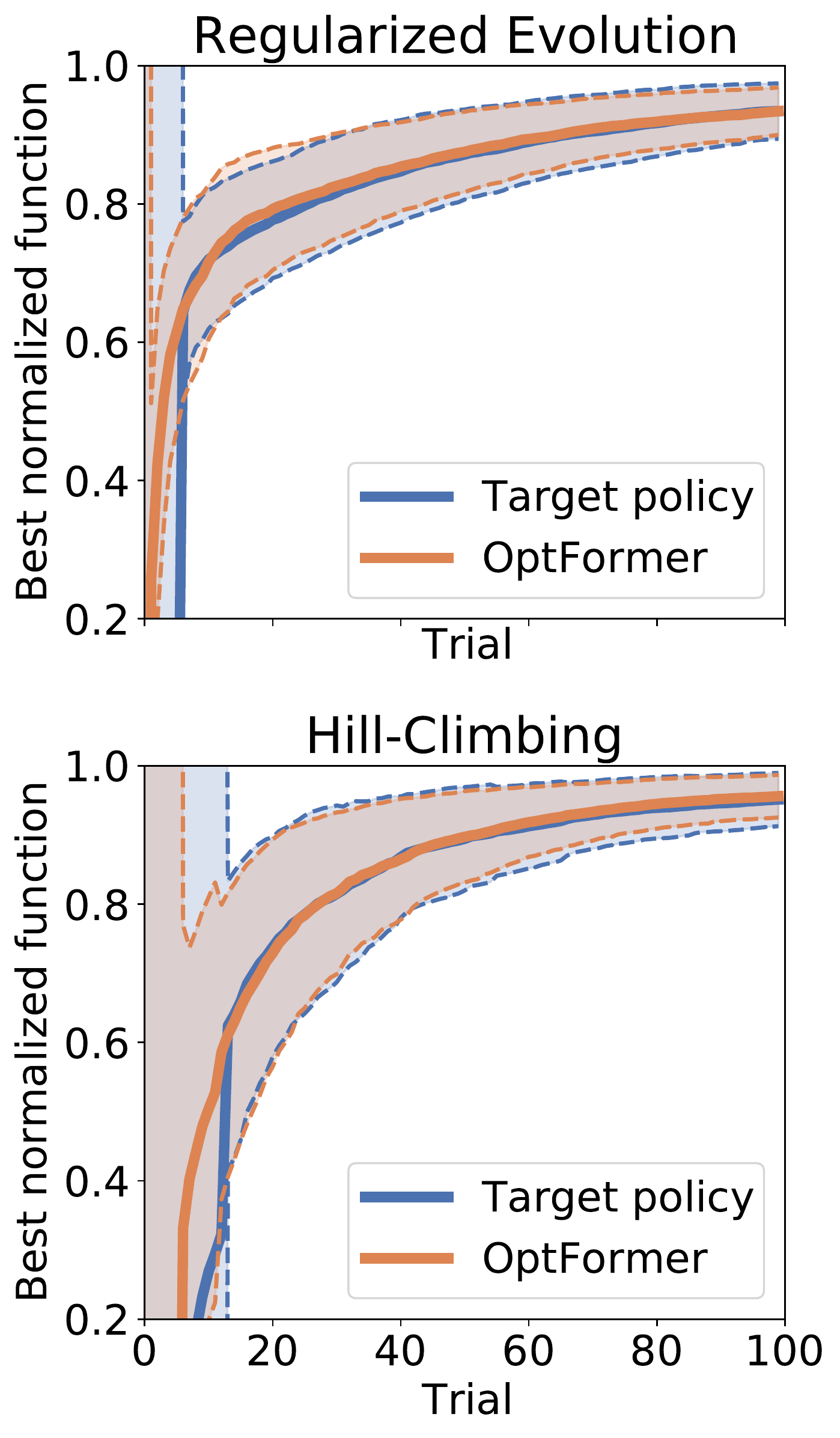}
        \caption{Optimization trajectory}
        \label{fig:bbob_y_curve_2alg}
    \end{subfigure}
    \hfill
    \begin{subfigure}[b]{0.32\textwidth}
        \centering
        \includegraphics[width=\textwidth]{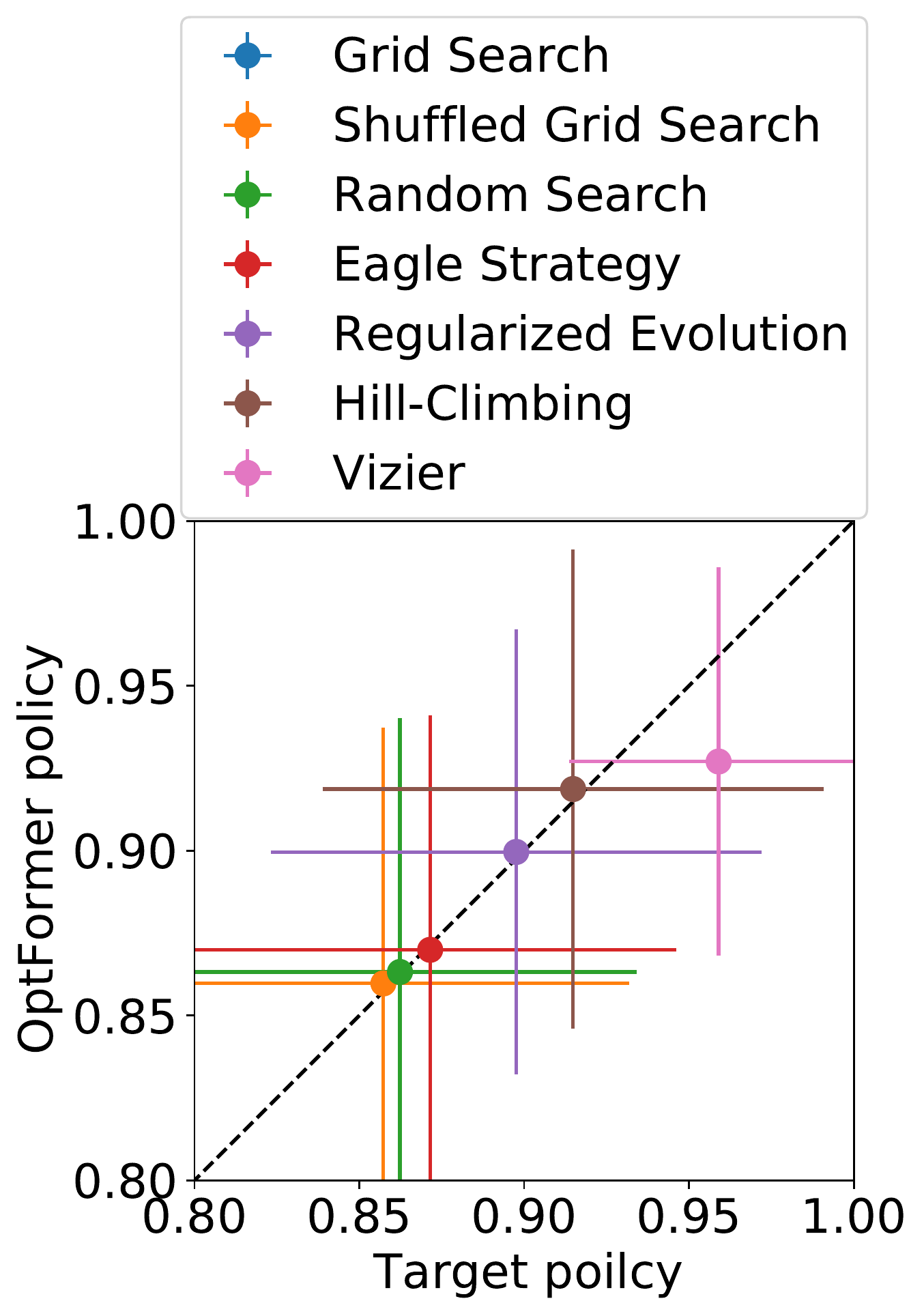}
        \caption{Average best function at trial 100 with standard deviation.} 
        \label{fig:bbob_y_at_trial100}
    \end{subfigure}
    \caption{Comparing the performance of different algorithms outputted by the \model conditioned on the corresponding algorithm's name.}
    \label{fig:bbob_imitate}
\vspace{-1.5em}
\end{figure}

We first evaluate how well the \model can learn the conditional distribution of parameter suggestions given by the behavior policies in the dataset, and how well it can imitate multiple algorithms.
As the algorithm's name is contained in the metadata $m$, we can modify the behaviour of the policy $\pi_{\text{prior}}(\vx_{t+1} | m, \vh_t)$ simply by altering this variable.
\cref{fig:bbob_imitate}\subref{fig:bbob_x_pred_at_trial_40_2alg} compares two different policies to the \model, when it is conditioned on the corresponding policy name. We observe a good match between the imitated algorithms and the \model (additional algorithms are shown in \cref{sec:more_exp_imitate}).

In \cref{fig:bbob_imitate}\subref{fig:bbob_y_curve_2alg} we run target policies on the \bbobdata dataset's test functions and compare the optimization trajectories of the algorithms and the \model. In \cref{fig:bbob_imitate}\subref{fig:bbob_y_at_trial100} we compare the average and standard deviation of the best normalized function values at trial 100. Our model imitates most algorithms very accurately in both the mean and variance except for the most complicated algorithm, Vizier, where $\pi_{\text{prior}}$ is slightly worse in the LUNACEK benchmark. We expand on this in \cref{sec:more_exp_imitate}. Because Vizier is the best performing HPO algorithm among all considered, the \model will imitate Vizier faithfully, although not perfectly, in the following experiments.

\subsection{Learning priors for hyperparameter response functions}
\label{sec:fun_prior}

In this section, we assess the \model's ability to learn the conditional distribution of the function value as a few-shot function regressor. Specifically, for every function in each test dataset, we repeatedly sample up to 200 random trials $(\vx_1, y_1, \dots \vx_{t}, y_{t}), t\leq 200$, and predict the conditional distribution $p(y_t|\vx_1, y_1, \dots, \vx_t)$. We compare with a GP model with output warping --- details provided in \cref{appendix:algorithms}. We report the log-predictive likelihood $\log p(y_t|\vx_t,\dots)$ in \cref{tab:fun_prior}.

\vspace{-0.05cm}
\begin{table}[h]
\begin{minipage}{0.54\textwidth}
    \caption{Log-predictive likelihood (with 1-std. standard error, higher is better ($\uparrow$)) and ECE (percentage of error, lower is better ($\downarrow$)) on \vizierdata and \hpobdata test sets.}
    \label{tab:fun_prior}
    \begin{tabular}{c|cc}
        \hline
                    & \multicolumn{2}{c}{Log-predictive likelihood $\uparrow$}\\
        Model       & \vizierdata & \hpobdata \\
        \hline
        GP          & $0.83 (0.06)$ & $4.03 (0.04)$ \\
        \model      & \textbf{2.12 (0.05)} & \textbf{6.16 (0.04)} \\
        \hline
        \hline
                    & \multicolumn{2}{c}{ECE (percent \%) $\downarrow$}\\
        Model       & \vizierdata & \hpobdata \\
        \hline
        GP          & 5.34 (0.06) & 2.39 (0.05)  \\
        \model      & \textbf{1.11 (0.02)} & \textbf{1.89 (0.01)} \\
        \hline
    \end{tabular}
\end{minipage}
\hfill
\vspace{.2cm}
\begin{minipage}{0.4\textwidth}
    \centering
    \vspace{.2cm}
    \includegraphics[width=\textwidth]{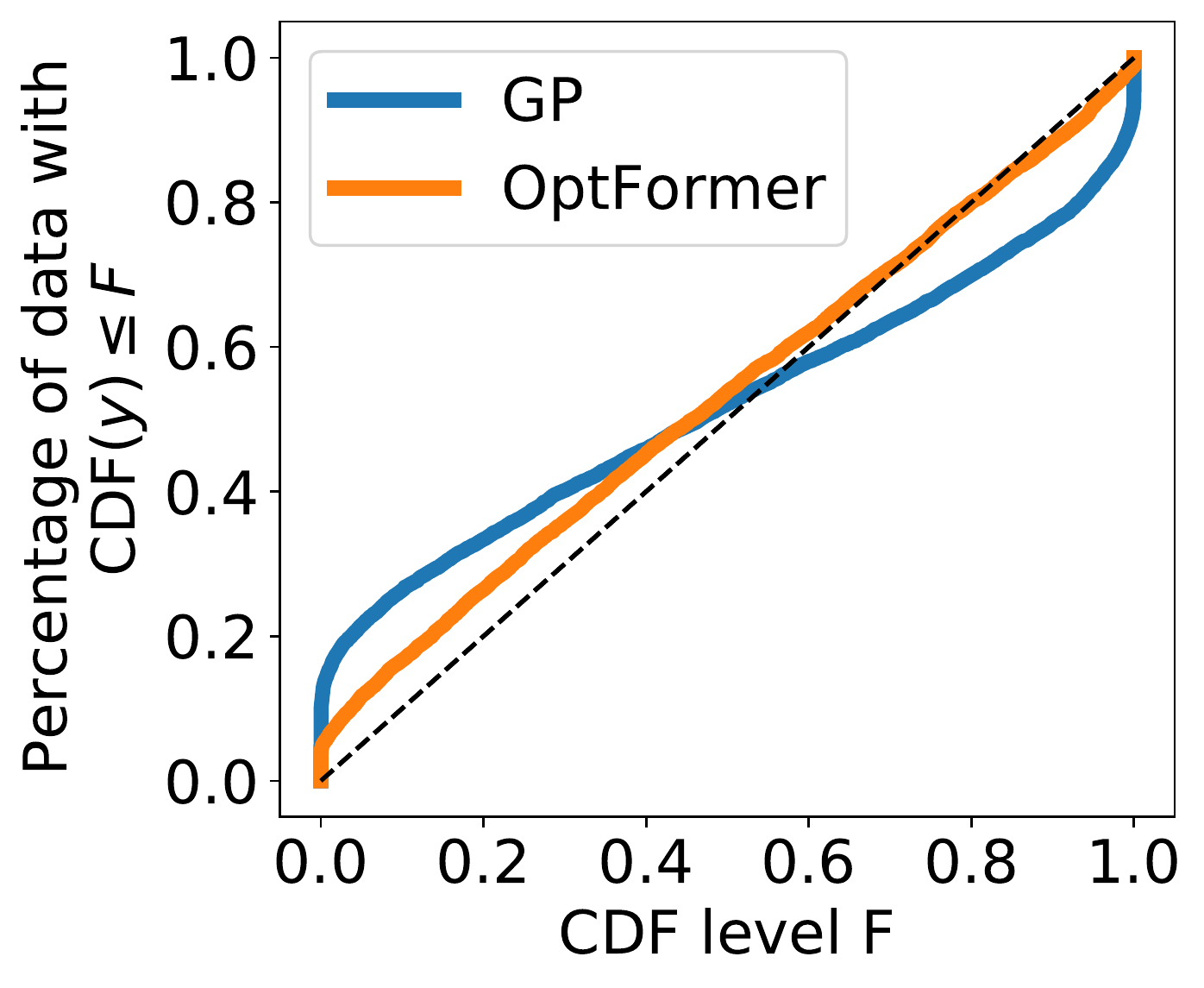}
    \captionof{figure}{Cumulative histogram of predicted $\mathrm{CDF(y)}$ on \vizierdata test set.}
    \label{fig:vizier_calibration}
\end{minipage}
\end{table}

As uncertainty estimation is important for HPO, we also evaluate how well the function predictive distribution is calibrated. When a predictive distribution $p_\theta(y|\dots)$ matches the true distribution, the estimated CDF $F(y)=\int_{-\infty}^y p_\theta(y'|\dots) dy'$ will be uniformly distributed. In \cref{fig:vizier_calibration}, we plot the cumulative histogram of $F(y)$ on \vizierdata test set and check the deviation from the diagonal line to assess goodness-of-fit as proposed by Rosenblatt~\citep{rosenblatt}. The \model has a smaller deviation than the GP almost across the entire range. We also compare calibration performance using the expected calibration error (ECE) \citep{naeini2015obtaining}. Readers are referred to \citep{naeini2015obtaining} and \cref{sec:more_exp_fun_prior} for a detailed explanation of ECE. We observe from \cref{tab:fun_prior} that the \model achieves better predictive likelihood and ECE than the GP on both datasets.

\subsection{Augmenting a prior policy with function prediction}
\label{sec:compare_policy}
We evaluate the \model as a hyperparameter optimization algorithm on two benchmarks, \vizierdata and \hpobdata. We compare our prior policy, the \model, and an augmented policy with Expected Improvement, the \model (EI), against standard HPO baselines, including Random Search, our implementation of GP-UCB, and the well-tuned Vizier service.
For \hpobdata, we also include the GP (not to be confused with our GP-UCB) and DGP (GP with deep kernel) baseline results provided by the original paper \citep{hpo-b}. Additionally, we include three recent transfer-learning methods based on multi-task GP models: ABLR~\cite{bishop2006pattern, perrone2018scalable}, FSBO~\cite{wistuba2020few}, and HyperBO~\cite{wang2018regret, wang2021hyperbo} (implementation details in \cref{appendix:algorithms}). Please note that all of these transfer learning methods require learning GPs on multiple tasks sharing the same search space. Therefore, none of them apply to the \vizierdata benchmark where every study has its own search space. 

\begin{figure}[t] 
    \centering
    \includegraphics[width=\textwidth]{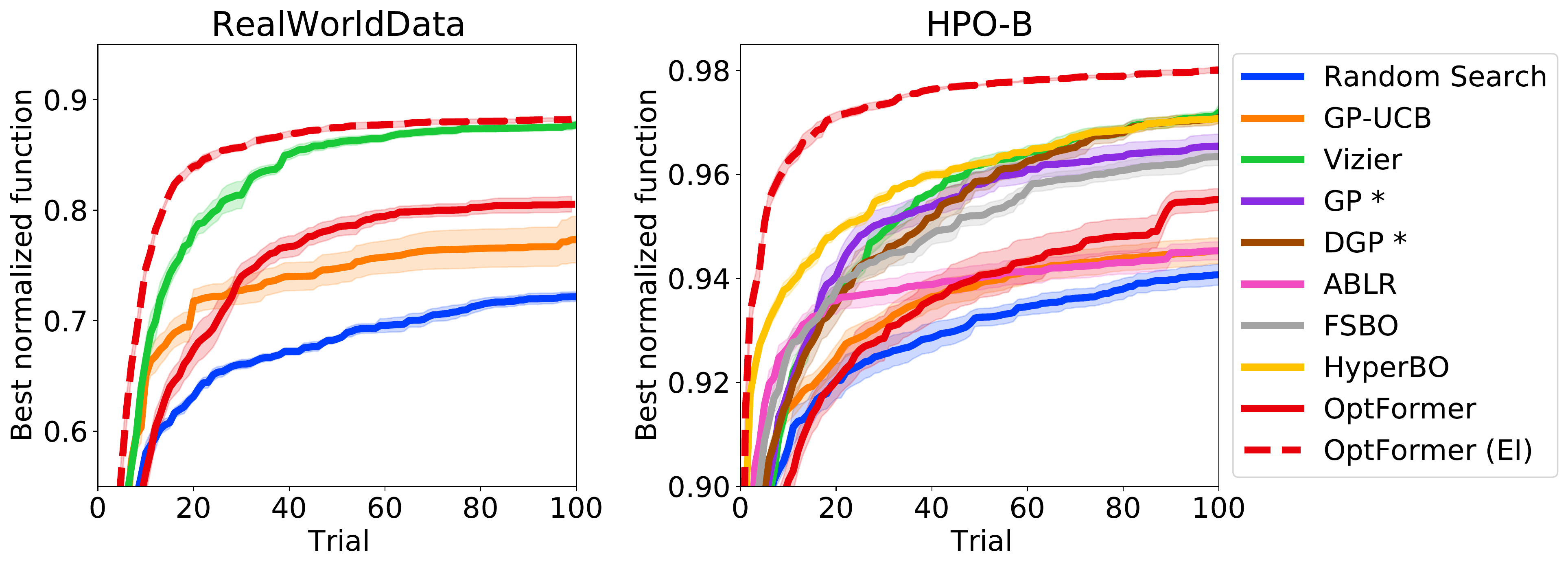}
    \caption{Higher is better. Best normalized function value averaged over 16 \vizierdata test functions (left) and over 86 \hpobdata test functions (right) with 1-std confidence interval from 5 runs. GP* and DGP* results are provided by \citep{hpo-b}. The transfer learning methods ABLR, FSBO and HyperBO cannot be applied to \vizierdata.}
    \label{fig:y_curve_per_step}
\end{figure}
\vspace{0.2cm}

We show the trajectory of the best normalized function value averaged over all functions from each benchmark in \cref{fig:y_curve_per_step}. While the prior policy returned by the \model does not perform as well as Vizier, it is comparable or slightly better than our GP-UCB baseline and ABLR.

The most significant improvement is achieved when we augment our prior policy with the Expected Improvement acquisition function. The resulting \model (EI) outperforms all baselines across the board on both benchmarks. This illustrates that the \model is able to learn the distribution of functions in the meta-training split and transfers to the meta-testing split.

It is worth noting that to run 100 trials for about half of the test functions, the required history token sequence is longer than the 1024-token length used in training, with the maximum length about twice the training horizon. The superior performance of the \model (EI) thus demonstrates its good generalization performance beyond the optimization horizon it is trained for.

\subsection{Ablations} 
\label{sec:ablation}
We provide further ablations on three important components for our policy:

\begin{figure}[t]
    \centering
    \begin{subfigure}[t]{0.45\textwidth}
        \centering
        \includegraphics[width=\textwidth]{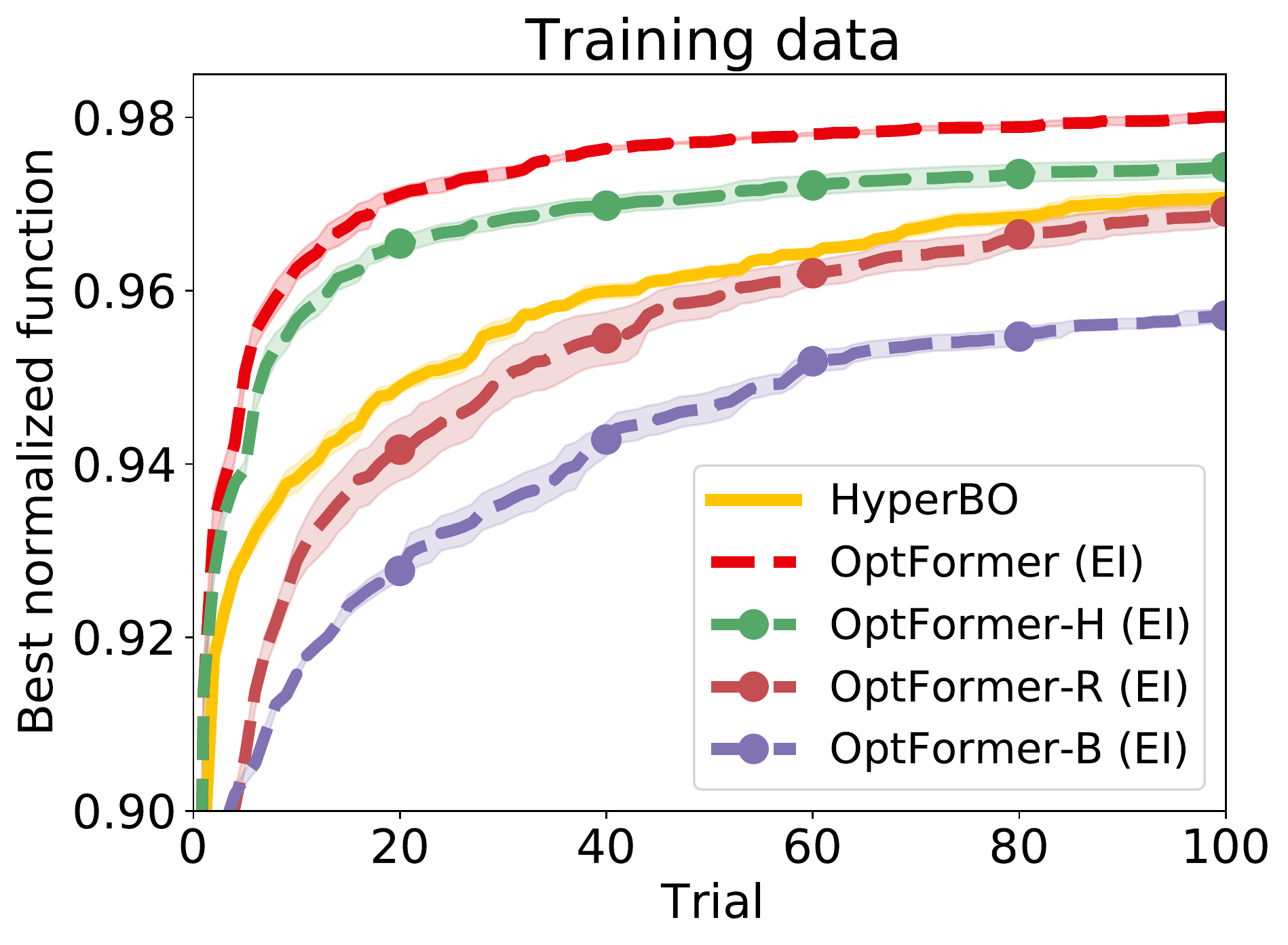}
        \caption{Ablation on training data.}
        \label{fig:ablation_data}
    \end{subfigure}
    \hfill
    \begin{subfigure}[t]{0.45\textwidth}
        \centering
        \includegraphics[width=\textwidth]{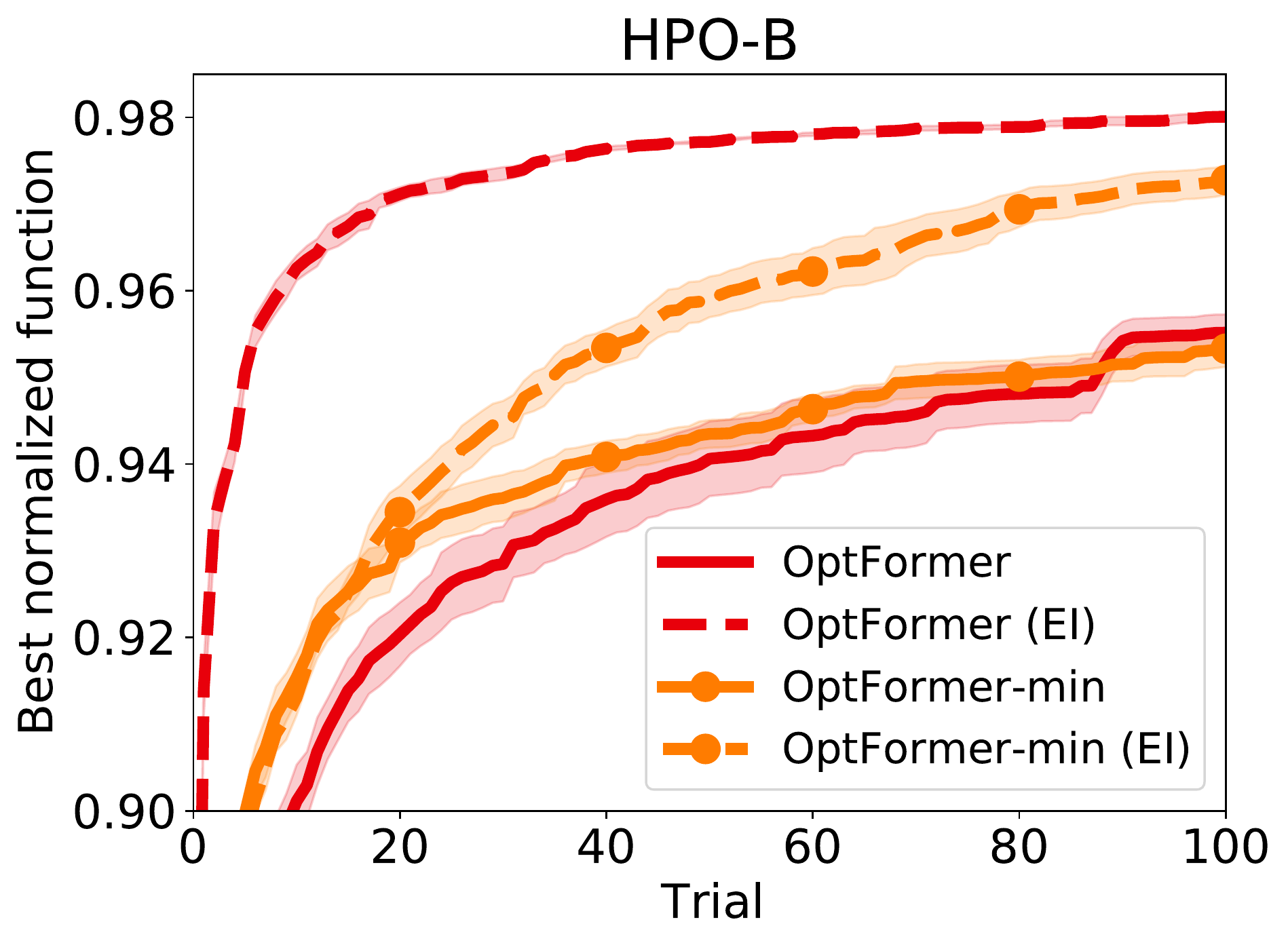}
        \caption{Ablation on metadata.}
        \label{fig:ablation_metadata}
    \end{subfigure}
    \\
    \begin{subfigure}[t]{0.45\textwidth}
        \centering
        \includegraphics[width=\textwidth]{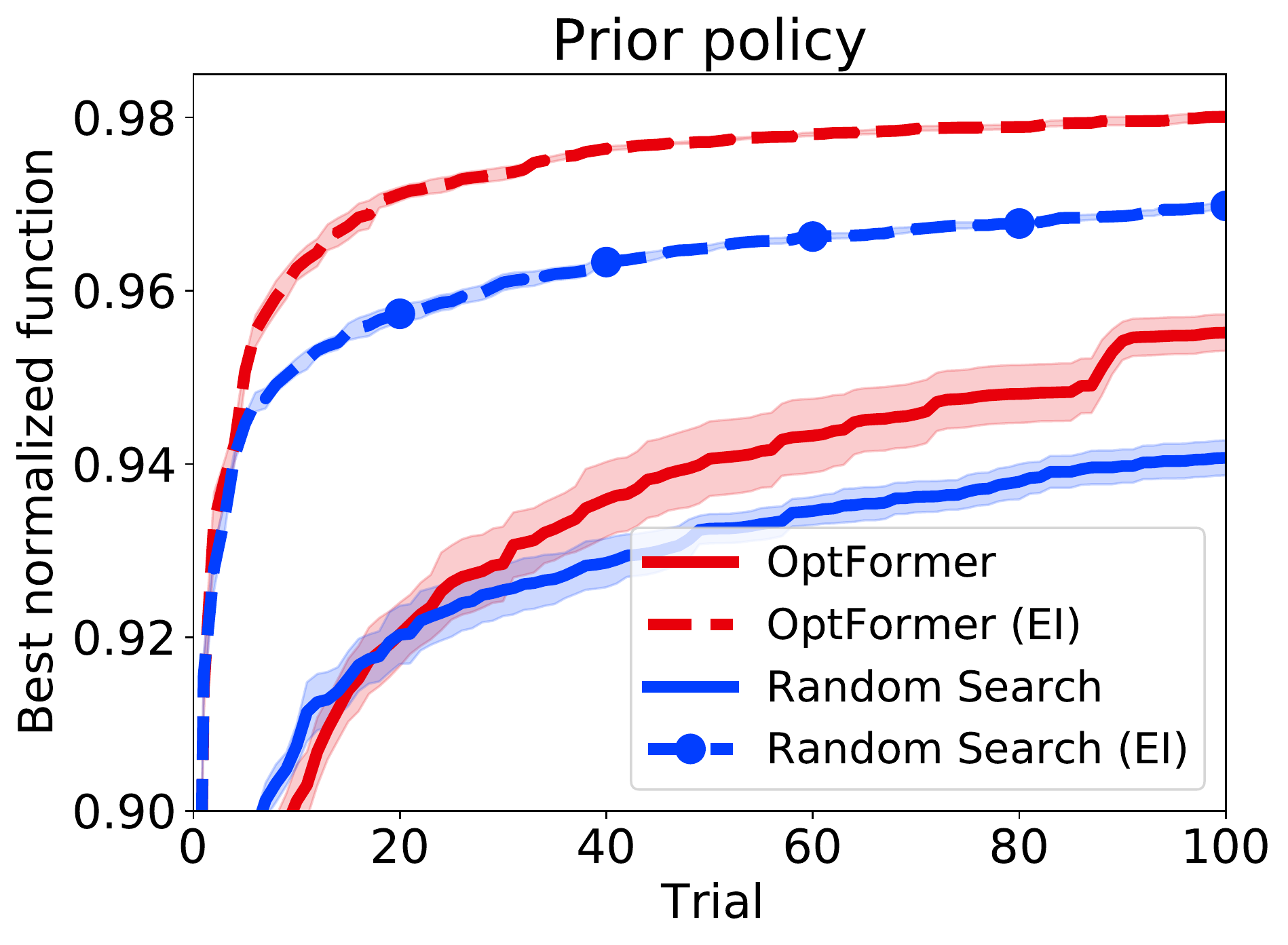}
        \caption{Ablation on the prior policy.}
        \label{fig:ablation_policy}
    \end{subfigure}
    \hfill
    \begin{subfigure}[t]{0.45\textwidth}
        \centering
        \includegraphics[width=\textwidth]{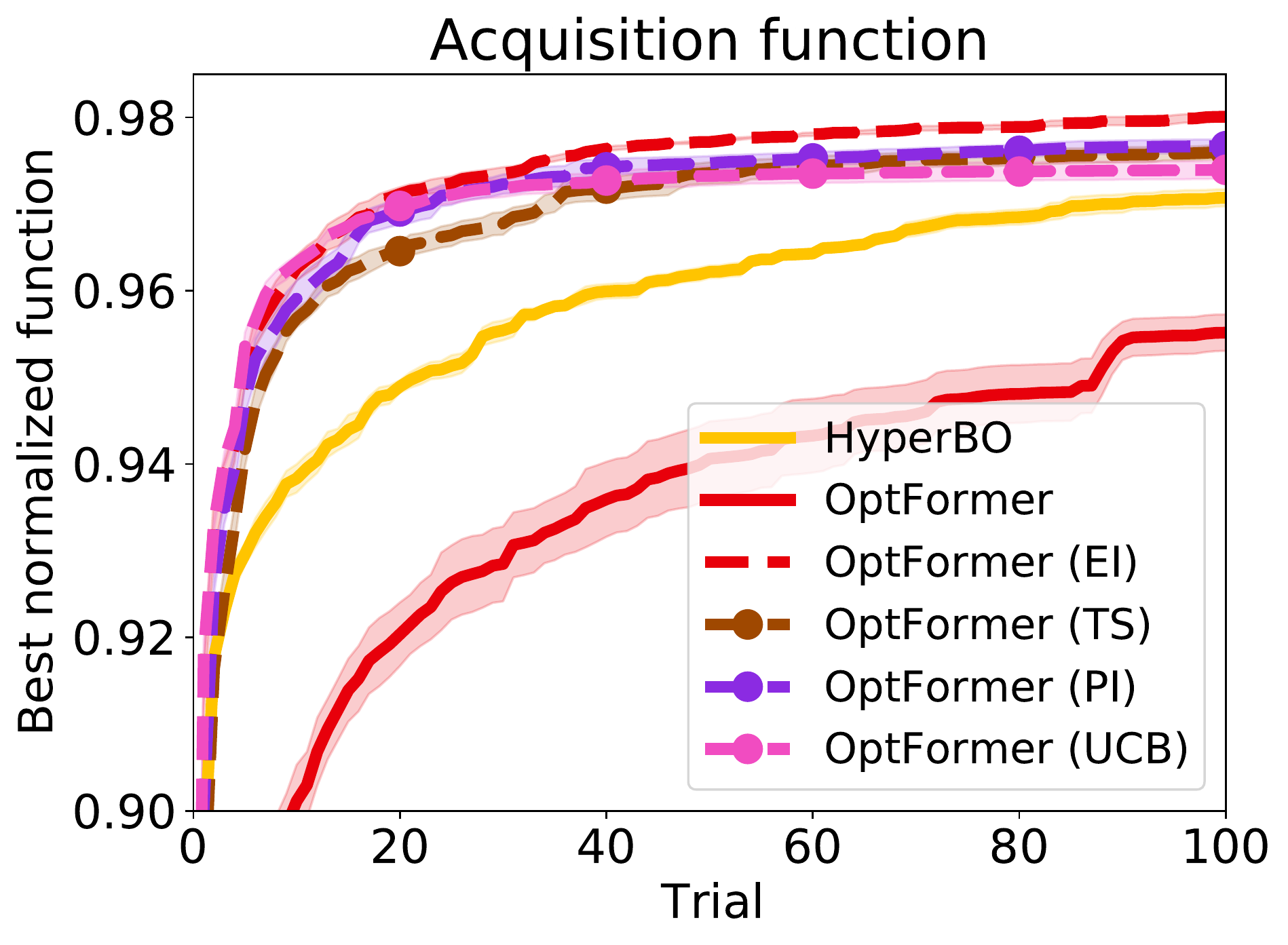}
        \caption{Ablation on the acquisition function.}
        \label{fig:ablation_aquisition}
    \end{subfigure}
    \caption{Best normalized function values averaged over \hpobdata test functions with 1-std confidence interval. Ablation curves are shown with $\bigcirc$ markers. (a) The more similar the training dataset, the better the transfer. Here, the suffix with "H", "R", "B" indicates training on \hpobdata, \vizierdata, and \bbobdata respectively. (b) Removing the majority of metadata hurts function prediction. (c) The prior policy improves performance with or without the Expected Improvement acquisition function. (d) All acquisition functions provide a significant improvement.}
    \label{fig:ablation}
\end{figure}

\paragraph{Training dataset.}
To understand the impact of the training datasets on the \model, we train three variants on individual datasets (\model-"R","H","B" respectively for \vizierdata, \hpobdata, \bbobdata) and study their transfer learning performances on \hpobdata. \cref{fig:ablation_data} verifies that training with in-domain data ("H") gives better performance than training over the more diverse across-domain \vizierdata HPO dataset ("R"), which is better than training over the synthetic \bbobdata data ("B"). Nonetheless, training on \vizierdata is enough to give comparable performance to the best transfer learning baseline at the end of 100 trials. Lastly, training on all of the datasets (\model) gives a further advantage over \modelh. This suggests that more data does not hurt the model's performance but rather may improve it, even if the extra data is out-of-domain.

\paragraph{Meta-data $m$.}
We have demonstrated how the \model's behavior can be controlled by the algorithm name in metadata $m$ in \cref{sec:imitate}. Here we study whether the \model learns to depend on other meta information. At inference time, we provide minimum information in $m$ (\model-min) by excluding all textual information and parameter value ranges. We only keep necessary information such as parameter types and algorithm names. \cref{fig:ablation_metadata} shows that the prior policy of \model-min performs comparably with the \model, partly due to the use of data augmentation (see \cref{sec:training_details}). The augmented policy \model-min (EI) (dashed orange) improves upon the prior policy but is significantly \textit{worse} than the full model, suggesting that the missing metadata impacts the model's predictions on function values.

\paragraph{Prior policy.}
\cref{sec:compare_policy} demonstrated the benefit of adding an acquisition function to the prior policy. A natural question is whether a good prior policy is needed at all. In \cref{fig:ablation_policy}, we replace the prior policy in the \model (EI) with random search (Random Search (EI), dashed blue line). While adding Expected Improvement still improves this random search policy's performance, the best method requires both a good prior policy and the acquisition function.

\paragraph{Choice of acquisition function.}
In \cref{fig:ablation_aquisition}, we compare the Expected Improvement (EI) with Thompson Sampling (TS), Probability of Improvement (PI), and Upper Confidence Bound (UCB) with a confidence level of 0.9. We observe that the prior policy is improved by all the acquisition functions. Particularly, \model (EI) is the best among all the choices though the difference is relatively small compared to the advantage over other baselines and \model prior policy. We provide additional analysis with results on both the \vizierdata and \hpobdata datasets, as well as other evaluation metrics in \cref{sec:more_exp_ablation}.

\section{Limitations and future extensions}
\label{sec:limitations}
We list a few limitations of this work and discuss some potential extensions. (1) We did not consider parameters that do not always apply or are subject to dynamic constraints depending on other parameter values. Such parameters are common in AutoML \citep{bergstra2013hyperopt} and NAS applications \citep{peng2020pyglove}. Our work can be extended to support these applications, by providing the conditional specifications as text in metadata $m$. (2) We also considered only sequential optimization with a batch size of 1. To support parallel suggestions, one could apply random masking to input function value observations to simulate scenarios with parallel pending trials \citep{chen2017learning}. (3) While we trained the Transformer to clone the behavior policy offline, there are extensive literature on offline RL \citep{levine2020tutorial} that could be applied here \citep{fujimoto2019off,kumar2020conservative,gulcehre2021regularized,upside-down-rl,decision_transformer,schrittwieser2021online}. One could also consider meta-training acquisition functions as in \citep{metalearn-acquisition} within the same model and online fine-tuning as in \citep{wistuba2020few,online_decision_transformer}. (4) We considered a single objective function, though multiple objectives can be easily included by outputting multiple function tokens in a trial. \edit{(5) The maximum sequence length is limited by the quadratic memory size requirement of a Transformer, which could be mitigated with more scalable architecture variants such as Performer \cite{choromanski2020rethinking}.}

\section{Conclusion}
We presented first step to learning a universal Transformer model for hyperparameter optimization from large scale datasets containing tuning experiments with vastly different search spaces and experiment descriptions. By training on a diverse set of synthetic and real-world tuning trajectories, we demonstrated the capacity of a single Transformer model to imitate 7 fundamentally different HPO policies, learn to make well calibrated few-shot function predictions, and provide competitive optimization performance on unseen test functions comparable with the existing, long-tried GP-based baselines. Many extensions are readily conceivable for future exploration.

\begin{ack}
We would like to thank Chris Dyer, Luke Metz, Kevin Murphy, Yannis Assael, and Esteban Real for providing valuable feedback during their reviews of this paper. We further thank Sebastian Pineda Arango for technical discussions on the \hpobdata benchmark and Christof Angermueller on biological benchmarks. In addition, we thank Daniel Golovin, Daiyi Peng, Yingjie Miao, Jack Parker-Holder, Jie Tan, Lucio Dery, and Aleksandra Faust for multiple useful conversations.
\end{ack}

\newpage

\bibliographystyle{unsrtnat}
\bibliography{references}

\newpage
\section*{Checklist}

\begin{enumerate}

\item For all authors...
\begin{enumerate}
  \item Do the main claims made in the abstract and introduction accurately reflect the paper's contributions and scope?
    \answerYes{Abstract and introduction point out the specific contributions in bullet points, which we have covered throughout or methodology and experiments.}
  \item Did you describe the limitations of your work?
    \answerYes{See \cref{sec:limitations}.}
  \item Did you discuss any potential negative societal impacts of your work?
    \answerYes{The main potential negative societal impacts of our work come from the inherent associated risks of using large Transformer models, such as bias. However, currently this does not apply to our work, as our data only contains numerical optimization data, and we are not using language models as of yet. Such a risk may grow however, if in future works, we e.g. use pretrained language models for warm-starting in our \model framework.}
  \item Have you read the ethics review guidelines and ensured that your paper conforms to them?
    \answerYes{The most relevant potential ethics concern is our use of \vizierdata. However, as we mention in \cref{appendix:data}, we anonymized the data to prevent users from being identified in the model.}
\end{enumerate}

\item If you are including theoretical results...
\begin{enumerate}
  \item Did you state the full set of assumptions of all theoretical results?
    \answerNA{Not applicable; no theoretical results.}
        \item Did you include complete proofs of all theoretical results?
    \answerNA{Not applicable; no theoretical results.}
\end{enumerate}

\item If you ran experiments...
\begin{enumerate}
  \item Did you include the code, data, and instructions needed to reproduce the main experimental results (either in the supplemental material or as a URL)?
    \answerYes{We included the link to the primary T5x codebase in \cref{sec:training_details}, along with relevant hyperparameters. But we do not include the training data. The datasets generated on public benchmarks, BBOB and HPO-B, can be reproduced by running publicly available HPO algorithms.}
  \item Did you specify all the training details (e.g., data splits, hyperparameters, how they were chosen)?
    \answerYes{These are specified in \cref{appendix:tokenization}, \cref{appendix:data}, and \cref{sec:training_details}.}
        \item Did you report error bars (e.g., with respect to the random seed after running experiments multiple times)?
    \answerYes{In experiments, we plotted error bars using 1 standard deviation over 5 runs.}
        \item Did you include the total amount of compute and the type of resources used (e.g., type of GPUs, internal cluster, or cloud provider)?
    \answerYes{See \cref{sec:training_details}.}
\end{enumerate}

\item If you are using existing assets (e.g., code, data, models) or curating/releasing new assets...
\begin{enumerate}
  \item If your work uses existing assets, did you cite the creators?
    \answerYes{We have cited \citep{ElHara2019COCOTL} for our use of \bbobdata dataset, \citep{hpo-b} for the \hpobdata dataset, and \citep{2020t5} for training T5 models.}
  \item Did you mention the license of the assets?
    \answerYes{For data, we mentioned the use of Apache 2.0 License when using the \hpobdata dataset.}
  \item Did you include any new assets either in the supplemental material or as a URL?
    \answerYes{We cited and linked the T5x codebase (see \cref{sec:training_details}) throughout this paper. We }
  \item Did you discuss whether and how consent was obtained from people whose data you're using/curating?
    \answerYes{The \bbobdata dataset is synthetic, and the \hpobdata dataset is public. For the \vizierdata dataset, users by default are agreeing to the condition that their tuning data may be used for research purposes. We have also ensured anonymity over the data by removing personally identifiable information.}
  \item Did you discuss whether the data you are using/curating contains personally identifiable information or offensive content?
    \answerYes{As discussed in \cref{appendix:data}, we anonymized user details from studies in the \vizierdata dataset.}
\end{enumerate}

\item If you used crowdsourcing or conducted research with human subjects...
\begin{enumerate}
  \item Did you include the full text of instructions given to participants and screenshots, if applicable?
    \answerNA{Not applicable.}
  \item Did you describe any potential participant risks, with links to Institutional Review Board (IRB) approvals, if applicable?
    \answerNA{Not applicable.}
  \item Did you include the estimated hourly wage paid to participants and the total amount spent on participant compensation?
    \answerNA{Not applicable.}
\end{enumerate}

\end{enumerate}

\clearpage
\appendix

\LARGE
\textsc{Appendix}
\normalsize

\section{Preprocessing and tokenization details}
\label{appendix:tokenization}

\subsection{Search space primitives}
\label{appendix:ss_primitives}
Below are the exact descriptions of the hyperparameter primitives used to define a given $\mathcal{X}^{(d)}$.

\begin{itemize}
\item \texttt{Double:} Specifies a continuous range of possible values in the closed interval $[x_\mathrm{min},x_\mathrm{max}]$ for some real values $x_\mathrm{min} \le x_\mathrm{max}$.
\item \texttt{Integer:} Specifies an integer range of possible values in $[x_\mathrm{min},x_\mathrm{max}] \in \mathbb{Z}$ for some integers $x_\mathrm{min} \le x_\mathrm{max}$.
\item \texttt{Discrete:} Specifies a finite, ordered set of values from $\mathbb{R}$.
\item \texttt{Categorical:} Specifies an unordered list of strings.
\end{itemize}

\subsection{Data preprocessing and tokenization}
\label{sec:details_preprocessing}
We list out the full set of preprocessing steps (from \cref{subsec:preprocessing}) below:
\begin{itemize}
    \item Omit parameter and metric names in all trials, remove redundant keywords (\texttt{"parameter"}, \texttt{"trial"}, etc.), \textbf{order trial parameters} according to those in metadata $m$, and add keywords (e.g., "name", "algorithm") and enumerating types (e.g. "DOUBLE") in the tokenizer vocabulary so that the original keywords are encoded into single tokens. 
    \begin{itemize}
        \item List of keywords: name, metric, goal, type, algorithm, min\_value, max\_value, scale\_type, categories.
        \item Enumerating values for the parameter type: DOUBLE, INTEGER, DISCRETE, CATEGORICAL.
        \item Enumerating values for the scale\_type: LINEAR, LOG.
    \end{itemize}
\item Insert short separator symbols, e.g. $\star$ between parameter/metrics in a trial, "|" between trials, and "\&" between experiment description and parameter configurations in metadata.
    \item Convert all values in history $h$ to single integers.
    \begin{itemize}
        \item Represent discrete and categorical parameters with their index in the set of values.
        \item Normalize float and integer parameter values in $x_t^{(d)}$ with their value range and the function values $y_t$ with their minimum and maximum seen values in the entire study. Then quantize the normalized values to an integer, e.g., \texttt{``0.12345"} $\rightarrow$ \texttt{"123"} with a quantization level of $Q=1000$. More formally, we apply the following transformation $q(\cdot)$:
        \begin{equation}
            q(z) = \mathrm{int} [z_{\mathrm{norm}} * Q], \text{ where } z_{\mathrm{norm}} = (z - z_\mathrm{min})/(z_\mathrm{max} - z_\mathrm{min})
            \label{eq:quantization_in_appx}
        \end{equation}
    \end{itemize}
\end{itemize}

The shortened text string is then converted to a sequence of tokens via the SentencePiece tokenizer \citep{kudo2018sentencepiece} with a vocabulary of 33000 words. Quantized numbers in $h$ are always converted into single tokens. As long as $Q$ is sufficiently large, there is no concern from the loss of precision over numerical quantizations, and thus the serialized study contains nearly the same amount of information as the original data.
For comparison, the naive tokenization for the example of \cref{tab:orig_study} with $t=100$ trials will produce 8221 tokens which can overload GPU memory, while our proposed tokenization will only produce 584 tokens, a 14x reduction.

\clearpage

\section{Algorithm and baseline details}
\label{appendix:algorithms}

\subsection{Dataset algorithms}

\paragraph{Grid Search:} \texttt{DOUBLE} parameters are first log-transformed if specified. They are then converted into \texttt{DISCRETE} parameters by discretizing their ranges into 100 equidistant points. Suggestions are outputted using lexicographic ordering from the cartesian product over all parameters' feasible points. The traversal order follows the alphabetical ordering of parameter names. That is, given two parameters \texttt{"Foo"} and \texttt{"Bar"}, both in [0,1] range, the sequence of trials looks like: \texttt{\{"Foo": 0, "Bar":0\} , \{"Foo": 0, "Bar":0.01\}, \ldots, \{"Foo": 0, "Bar":1\}, \{"Foo": 0.01, "Bar":0\}, \{"Foo": 0.01, "Bar":0.01\}, \ldots}.

\paragraph{Shuffled Grid Search:} Shuffled grid search is the same as Grid Search in how it handles \verb+DOUBLE+ parameters. Instead of traversing the grid in a deterministic order, it selects without replacement a random point from the grid at each iteration.

\paragraph{Regularized Evolution \citep{real2019regularized}:}  In summary, this algorithm at every iteration randomly selects a tournament subset from the current population, and mutates the argmax member of the tournament. When inserting a new trial, the oldest trial will be removed. We use a population size of 25 and tournament size of 5. The mutation operation uniformly at random selects one of the parameters $x^{(r)}$ from $\vx$, and mutates $x^{(r)}$ based on the following: for \texttt{DOUBLE, INTEGER}, the new value is uniformly sampled from $\left[x^{(r)}_\mathrm{min}, x^{(r)}_\mathrm{max}\right]$, while for \texttt{DISCRETE, CATEGORICAL}, the new value is uniformly sampled from the feasible list.

\paragraph{Hill Climbing:} This is a naive implementation, where at every iteration $t$, the current $\vx_\mathrm{pivot}$ is mutated (using the same operation as Regularized Evolution) to $\vx_\mathrm{mutated}$, and evaluated. If $f(\vx_\mathrm{mutated}) > f(\vx_\mathrm{pivot})$, then we reassign $\vx_\mathrm{pivot}$ to be the mutated $\vx_\mathrm{mutated}$. An extension of this method can be "batched", as seen in \citep{golovin2019gradientless}, although we not include this for the sake of clarity and presentation.

\paragraph{Eagle Strategy \cite{yang2010eagle}:} Eagle strategy is a metaheuristics algorithm that is a slight variation of Particle Swarm Optimization \cite{kennedy1995particle}. 

The algorithm is originally formulated for continuous search spaces only. The reason is that it involves a subroutine (\texttt{move} step) where we take a convex combination of a particle (called \emph{firefly} in \cite{yang2010eagle}) and another particle that has a better objective value. Mathematically, given two particle vectors $\vx$ and $\vx'$ and the coefficient $c \in [0,1]$, the \texttt{move} step generates $c \vx + (1-c) \vx'$.

The algorithm is extended to support \verb+DISCRETE+ and \verb+CATEGORICAL+ parameters by applying a separate \verb+move+ operation for each non-continuous dimension $d$:
$$
\texttt{move}(x^{(d)}, x'^{(d)}, c, \alpha) =  
\begin{cases}
x^{(d)} & \text{with probability } (1-\alpha) c \\
x'^{(d)} &  \text{with probability } (1-\alpha) (1-c) \\
\text{random value} & \text{with probability } \alpha
\end{cases}
$$
where $\alpha$ is a small perturbation coefficient that decreases in the dimension of the search space.

\paragraph{Vizier \citep{vizier}:} Vizier's default algorithm is available via Google Cloud as Vertex Vizier. We have contacted the authors of the algorithm and received the the following details on its implementation. 

In summary, the algorithm uses a particular implementation of GP-UCB with trust regions. The GP regressor model consists of the following:
\begin{itemize}
\item $\alpha \sim \mathrm{TruncatedLogNormal}$ controls the amplitude of Matern5/2 kernel.
\item $\lambda_i \sim \mathrm{TruncatedLogNormal}$ (i.i.d. for each dimension $i$) controls the length scale for the $i$-th dimension.
\item $\sigma \sim \mathrm{TruncatedLogNormal}$ controls the Gaussian noise.
\item $z \sim \mathrm{Normal}(0, \sigma)$ is the observation noise.
\item $f \sim \mathrm{GP}(\lambda, \alpha)$ is the function.
\item $y(x) \sim f(x) + z$ is the noisy function.
\end{itemize}
where the prior distribution parameters are chosen depending on the user's estimate of the observation noise.

The algorithm then uses gradient descent with line search for step sizes to obtain the MAP estimate of $\alpha, \lambda$ and $\sigma$. Furthermore, the algorithm uses a variation of Eagle Strategy (explained above) to optimize the UCB acquisition function with coefficient of $1.8$. In order to prevent overexploration that may result from the large UCB coefficient, the algorithm optimizes acquisition functions inside trust region. The trust region is the union of $L_\infty$-norm balls around explored points. The radius of the $L_\infty$-norm ball grows in the number of explored points. The algorithm also starts at the center of the search space (unless user specifies an alternative initial batch).

\paragraph{GP-UCB:} 
It is the same as Vizier's GP-UCB, except for the model definition. We used the model definition from the github repository of the authors of "Heteroscedastic and Evolutionary Bayesian Optimisation solver" (HEBO) \citep{cowen2020empirical_hebo}, the winner of 2020 Blackbox Optimization challenge \citep{turner2021bayesian_bbochallenge}. It is worth noting that HEBO uses multi-dimensional acquisition functions derived from the GP model. The priors over hyperparameters are thus not tuned to optimize the performance of GP-UCB algorithm, which explains its suboptimal performance.

%

\subsection{Gaussian Process for uncertainty estimation}

We use the same GP model as GP-UCB.

When comparing the function prediction performance with the \model,
we choose $[y_\mathrm{min}, y_\mathrm{max}]$ to normalize function value token based on the range of observed value in the sampled sequence $(\vx_1, y_1, \dots \vx_{t}, y_{t})$, and therefore the real value of $y_t$ always resides in the prediction support of the \model.

To compensate for the fact that GP's distribution is wider than the real support used by the Transformer, we truncate the GP's prediction into $[y_\mathrm{min}, y_\mathrm{max}]$ for a fairer comparison.

\subsection{Transfer learning baselines}
We use the following methods as transfer-learning baselines for the HPO-B dataset from \cref{sec:compare_policy}:

\paragraph{ABLR \cite{bishop2006pattern, perrone2018scalable}:} BO with multi-task adaptive Bayesian linear regression. Our implementation of ABLR is equivalent to a GP with 0 mean and a dot-product kernel with learned basis functions. We use a neural net (NN) with $(128,128)$ hidden layers and tanh activation as the basis functions. We then train ABLR by optimizing the negative log likelihood (NLL) over NN weights $\theta$ as well covariance matrix $SS\T$ and bias parameters $\delta^2$ that define the dot-product kernel $k$, i.e. 
\begin{equation}
k(x, x') = \phi_\vartheta(\vx)\T SS\T\phi_{\vartheta}(\vx') + \delta^2,
\end{equation}
where matrix $S \in \R^{128\times 256}$, basis function $\phi_\theta$ is parameterized by NN weights $\vartheta$ and $\delta\in \R$. 

\paragraph{FSBO \cite{wistuba2020few}:} Bare-bone few-shot BO. We did not include data-driven initialization due to lack of reproducing details. Following~\cite{wistuba2020few}, our implementation of FSBO is equivalent to BO using a GP with 0 mean and a squared-exponential kernel on top of a NN with $(128,128)$ hidden layers and tanh activation functions. We train the NN weights as well as the parameters in the squared-exponential kernel.

\paragraph{HyperBO \cite{wang2018regret, wang2021hyperbo}:} BO with pre-trained GPs. Following~\cite{wang2021hyperbo}, we pre-train a GP with Mat\'ern32 kernel on top of a NN with one hidden layer of width $2\times D$ and tanh activation functions. Here $D$ is the input dimension of the search space.

For training, we use the Adam optimizer with learning rate $0.001$ and batch size $50$ for all the transfer-learning baselines. 
Notice that these transfer-learning methods require ``pre-training'' a GP on the same search space. We sample $10000$ random data points on each \hpobdata surrogate functions from each search space. We train a separate GP for each search space.

\clearpage

\section{Data details}
\label{appendix:data}

\subsection{Dataset details}
\paragraph{\vizierdata dataset:} The \vizierdata dataset contains a total of 750K studies collected from Google Vizier users over a span of 5 years (starting from 2017), and each study has a variable number of trials. Since some user studies can potentially have an excessive number of trials (e.g. 10K+), for all studies we only consider the first 300 trials for experiments. Since the dataset also includes Google employee usernames, we made sure to anonymize every study first.

We split the dataset in temporal order to avoid information leak, use most studies for training, and select 16 studies generated by a different set of users for testing. \edit{All training studies were generated before Feb 2020. The test studies were created by users who started to use the hyperparameter tuning service after that date.} To bootstrap these studies into actual functions to be evaluated, we fit a GP for each study and output the function value as the GP's median function (due to the use of output warping).


\paragraph{\hpobdata dataset:} For HPO-B dataset, a tuning task is identified with a (search space id, dataset id) pair, which refers to tuning the hyperparameters defined in a search space for some machine learning model trained on a dataset.
we use the \texttt{"v3-augmented"} meta-training/validation/test splits that includes all the 16 test search spaces as well as less frequent search spaces in the meta-training split. There are uniquely 1730, 91, and 86 tasks for training, validation and testing respectively. For every tuning task, \citep{hpo-b} fits an XGBoost model to the trial data of every tuning task as the objective function.

Similar to the \bbobdata dataset, we generate 10M, 500K studies for training and validation respectively, along with the same set of controlled algorithms. For each of the test tuning task, we run 5 optimizations each with a different initial set of observations provided in \citep{hpo-b}.

The \hpobdata uses the Apache 2.0 open-source license.

\paragraph{\bbobdata dataset:} The \bbobdata dataset contains a total of 10M studies for training, each containing exactly 300 trials. An additional 500K studies (using different randomization seeds) are used for validation. While the number of studies can be freely generated and effectively unlimited, we found that 10M studies were sufficient for the Transformer to train properly. 

The functions we use for data are from \citep{ElHara2019COCOTL}, and consist of separable functions (\texttt{Sphere, Ellipsoid Separable, Rastrigin Separable, Bueche Rastrigin, \{Linear Slope\}}), moderately conditioned, potentially multi-modal functions (\texttt{Attractive Sector, Step Elllipsoid, \{Rosenbrock Rotated\}}), ill-conditioned functions (\texttt{Discus, Bent Cigar, Sharp Ridge, \{Sum of Powers\}}), multi-modal functions (\texttt{Weierstrass, Schaffers F7, Schaffers F7 Illconditioned, \{Greiwank Rosenbrock\}}), and functions with weak global structures (\texttt{Schwefel, Gallagher 21, Gallagher 101, Katsuura, \{Lunacek\}}). The functions noted with the extra "\{\}" are for testing and excluded from the training data. We apply significant randomization over the functions for both the training dataset and test-time evaluation. In order, we randomize the following:

\begin{itemize}
\item Function dimension $D$, which is uniformly selected from a range. For training data generation, this range is $[1, 20]$.
\item Orthonormal rotation matrix $\Gamma$, which is applied to the input first, i.e. producing a new function $f'(\vx) = f(\Gamma \vx)$.
\item Shift vector $\vx_{shift}$ which is also applied to the input first, i.e. producing a new function $f'(x) = f(x- \vx_{shift})$, where $\vx_{shift}$ has all of its coordinate-wise entries sampled from $[-4,4]$, while the domain is $[-5,5]$.
\item Discretization, in which the parameter space $\mathcal{X}^{(d)}$ is uniformly at random chosen to be either a \texttt{DOUBLE}, \texttt{DISCRETE}, \texttt{CATEGORICAL} parameter. The \texttt{DOUBLE} parameter "discretization" is actually a no-op, as it allows the original continuous space $\mathcal{X}^{(d)} \subset \mathbb{R}$. Otherwise, a number $L$ of feasible points is uniformly selected from the range $[2,8]$, and used to divide the original $[-5,5]$ range into $L$ equally-spaced points. If \texttt{DISCRETE} was chosen, then the ordering of the grid points is preserved, otherwise if \texttt{CATEGORICAL} was chosen, then all of the gridpoints become effectively unordered strings.
\item Noise Type, in which one of 10 noise settings (including no noise) is uniformly chosen. Noise consists of either Gaussian (multiplier sampled from a random Gaussian of varying scale is applied), Uniform (multiplier sampled from uniform distribution of varying scale is applied), or Cauchy (additive noise which only occurs at a probabilistic frequency, with a varying fixed strength is applied).
\end{itemize}

For evaluation, we randomly sample 100 configurations for each of the five test functions, resulting in 500 optimization trajectories in total.

For \bbobdata, as all parameters are named as \texttt{"x\_i"} with $i\in[0,D)$ and always have value range in $[-5,5]$, significantly different from the other two datasets, we omit their parameter names and value in the metadata $m$ and only keep parameter type information.

\begin{table}[tbp]
    \centering
    \caption{Example of studies in \vizierdata (left), \bbobdata (middle) and \hpobdata (right).}
    \label{tab:dataset_study_example}
    \hspace*{-1.4cm}
    {\scriptsize
    \begin{tabular}{c|c|c}
    \toprule
    \begin{minipage}[t]{2in}
    \begin{Verbatim}[commandchars=\\\{\}]
\textcolor{blue}{"name": "gan1d 500 iters - }
\textcolor{blue}{"2022-05-18"}
\textcolor{blue}{"parameter": \{}
\textcolor{blue}{  "name": "learning_rate",}
\textcolor{blue}{  "min_value": 1e-06,}
\textcolor{blue}{  "max_value": 0.01,}
\textcolor{blue}{  "type": "DOUBLE",}
\textcolor{blue}{  "scale_type": "LOG",}
\textcolor{blue}{\}}
\textcolor{blue}{"parameter": \{}
\textcolor{blue}{  "name": "modifier",}
\textcolor{blue}{  "min_value": 0.1,}
\textcolor{blue}{  "max_value": 1000000.0,}
\textcolor{blue}{  "type": "DOUBLE",}
\textcolor{blue}{  "scale_type": "LOG",}
\textcolor{blue}{\}}
\textcolor{blue}{"parameter": \{}
\textcolor{blue}{  "name": "weight_init_std",}
\textcolor{blue}{  "min_value": 0.01,}
\textcolor{blue}{  "max_value": 2.0,}
\textcolor{blue}{  "type": "DOUBLE",}
\textcolor{blue}{\}}
\textcolor{blue}{"parameter": \{}
\textcolor{blue}{  "name": "optimizer",}
\textcolor{blue}{  "type": "CATEGORICAL",}
\textcolor{blue}{  "categories": "sgd",}
\textcolor{blue}{  "categories": "adam",}
\textcolor{blue}{  "categories": "rmsprop",}
\textcolor{blue}{\}}
\textcolor{blue}{"goal": "MINIMIZE",}
\textcolor{blue}{"max_num_trials": 500,}
\textcolor{blue}{"metric": "",}
\textcolor{blue}{"observation_noise": "HIGH",}
\textcolor{purple}{"trial": \{}
\textcolor{purple}{ "parameter": \{}
\textcolor{purple}{   "learning_rate": 0.0001,}
\textcolor{purple}{   "modifier": }
\textcolor{purple}{         316.2277660168381,}
\textcolor{purple}{   "optimizer": "sgd",}
\textcolor{purple}{   "weight_init_std": 1.005,}
\textcolor{purple}{ \}}
\textcolor{purple}{ "metric": \{}
\textcolor{purple}{   "": -0.946908021738347,}
\textcolor{purple}{ \}}
\textcolor{purple}{\}}
\textcolor{purple}{"trial": \{}
\textcolor{purple}{ "parameter": \{}
\textcolor{purple}{   "learning_rate": 0.000504,}
\textcolor{purple}{   "modifier": }
\textcolor{purple}{          12.346786652749216,}
\textcolor{purple}{   "optimizer": "rmsprop",}
\textcolor{purple}{   "weight_init_std": }
\textcolor{purple}{          1.2192566347109868,}
\textcolor{purple}{ \}}
\textcolor{purple}{ "metric": \{}
\textcolor{purple}{   "": -1.5144472008077585,}
\textcolor{purple}{ \}}
\textcolor{purple}{\}}
\textcolor{purple}{...}
    \end{Verbatim}
    \end{minipage}
    
    &
    
    \begin{minipage}[t]{2in}
    \begin{Verbatim}[commandchars=\\\{\}]
\textcolor{blue}{"name": "SCHAFFERS_F7",}
\textcolor{blue}{"algorithm": "gp",}
\textcolor{blue}{"parameter": \{}
\textcolor{blue}{  "name": "x0",}
\textcolor{blue}{  "type": "CATEGORICAL",}
\textcolor{blue}{  "categories": ["0.0", "5.0",}
\textcolor{blue}{                 "-5.0"],}
\textcolor{blue}{\},}
\textcolor{blue}{"parameter": \{}
\textcolor{blue}{  "name": "x1",}
\textcolor{blue}{  "min_value": -5.0,}
\textcolor{blue}{  "max_value": 5.0,}
\textcolor{blue}{  "type": DOUBLE,}
\textcolor{blue}{  "scale_type": UNIT_LINEAR_SCALE,}

\textcolor{blue}{\},}
\textcolor{blue}{"parameter": \{}
\textcolor{blue}{  "name": "x2",}
\textcolor{blue}{  "min_value": -5.0,}
\textcolor{blue}{  "max_value": 5.0,}
\textcolor{blue}{  "type": DOUBLE,}
\textcolor{blue}{  "scale_type": UNIT_LINEAR_SCALE,}

\textcolor{blue}{\},}
\textcolor{blue}{"parameter": \{}
\textcolor{blue}{  "name": "x3",}
\textcolor{blue}{  "type": DISCRETE,}
\textcolor{blue}{  "values": [-5.0, 5.0],}
\textcolor{blue}{\},}
\textcolor{blue}{"parameter": \{}
\textcolor{blue}{  "name": "x4",}
\textcolor{blue}{  "type": CATEGORICAL,}
\textcolor{blue}{  "categories": ["5.0",}
\textcolor{blue}{      "-1.66666666667",}
\textcolor{blue}{                 "-5.0",}
\textcolor{blue}{      "1.666666666667"],}
\textcolor{blue}{\},}
\textcolor{blue}{"parameter": \{}
\textcolor{blue}{  "name": "x5",}
\textcolor{blue}{  "min_value": -5.0,}
\textcolor{blue}{  "max_value": 5.0,}
\textcolor{blue}{  "type": DOUBLE,}
\textcolor{blue}{  "scale_type": UNIT_LINEAR_SCALE,}

\textcolor{blue}{\}}
\textcolor{blue}{"metric": "",}
\textcolor{blue}{"goal": MAXIMIZE,}
\textcolor{blue}{"observation_noise": HIGH}
\textcolor{purple}{"trial": \{}
\textcolor{purple}{ "parameter": \{}
\textcolor{purple}{   "x0": "0.0",}
\textcolor{purple}{   "x1": 0.0,}
\textcolor{purple}{   "x2": 0.0,}
\textcolor{purple}{   "x3": 5.0,}
\textcolor{purple}{   "x4": "-5.0",}
\textcolor{purple}{   "x5": 0.0,}
\textcolor{purple}{ \}}
\textcolor{purple}{ "metric": \{}
\textcolor{purple}{   "": -334.4782223514127,}
\textcolor{purple}{ \}}
\textcolor{purple}{\}}
\textcolor{purple}{"trial": \{}
\textcolor{purple}{ "parameter": \{}
\textcolor{purple}{   "x0": "5.0",}
\textcolor{purple}{   "x1": -1.9867479768748013,}
\textcolor{purple}{   "x2": -1.7665621302793095,}
\textcolor{purple}{   "x3": -5.0,}
\textcolor{purple}{   "x4": "1.666666666666667",}
\textcolor{purple}{   "x5": -1.7634306558106605,}
\textcolor{purple}{ \}}
\textcolor{purple}{ "metric": \{}
\textcolor{purple}{   "": -323.84900527589326,}
\textcolor{purple}{ \}}
\textcolor{purple}{\}}
\textcolor{purple}{...}
    \end{Verbatim}
    \end{minipage}
    
    &
    
    \begin{minipage}[t]{2.1in}
    \begin{Verbatim}[commandchars=\\\{\}]
\textcolor{blue}{"name": "5859_145853",}
\textcolor{blue}{"algorithm": "GP UCB",}
\textcolor{blue}{"parameter": \{}
\textcolor{blue}{  "name": "minsplit",}
\textcolor{blue}{  "max_value": 60.0,}
\textcolor{blue}{  "type": "DOUBLE",}
\textcolor{blue}{  "scale_type": "LINEAR",}
\textcolor{blue}{\}}
\textcolor{blue}{"parameter": \{,}
\textcolor{blue}{  "name": "minsplit.na",}
\textcolor{blue}{  "max_value": 1.0,}
\textcolor{blue}{  "type": "DOUBLE",}
\textcolor{blue}{\}}
\textcolor{blue}{"parameter": \{}
\textcolor{blue}{  "name": "minbucket",}
\textcolor{blue}{  "min_value": 1.0,}
\textcolor{blue}{  "max_value": 60.0,}
\textcolor{blue}{  "type": "DOUBLE",}
\textcolor{blue}{  "scale_type": "LINEAR",}
\textcolor{blue}{\}}
\textcolor{blue}{"parameter": \{}
\textcolor{blue}{  "name": "cp",}
\textcolor{blue}{  "min_value": 0.000100788830221,}
\textcolor{blue}{  "max_value": 1.000092678873241,}
\textcolor{blue}{  "type": "DOUBLE",}
\textcolor{blue}{  "scale_type": "LOG",}
\textcolor{blue}{\}}
\textcolor{blue}{"parameter": \{}
\textcolor{blue}{  "name": "maxdepth",}
\textcolor{blue}{  "max_value": 29.0,}
\textcolor{blue}{  "type": "DOUBLE",}
\textcolor{blue}{  "scale_type": "LINEAR",}
\textcolor{blue}{\}}
\textcolor{blue}{"parameter": \{}
\textcolor{blue}{  "name": "maxdepth.na",}
\textcolor{blue}{  "max_value": 1.0,}
\textcolor{blue}{  "type": "DOUBLE",}
\textcolor{blue}{\}}
\textcolor{blue}{"observation_noise": AUTOMATIC,}
\textcolor{blue}{"metric": "objective_value",}
\textcolor{blue}{"goal": "MAXIMIZE"}
\textcolor{purple}{"trial": \{}
\textcolor{purple}{ "parameter": \{}
\textcolor{purple}{   "minsplit": 4.0,}
\textcolor{purple}{   "minsplit.na": 0.0,}
\textcolor{purple}{   "minbucket": 18.0,}
\textcolor{purple}{   "cp": 0.7342895964927976,}
\textcolor{purple}{   "maxdepth": 3.0,}
\textcolor{purple}{   "maxdepth.na": 0.0,}
\textcolor{purple}{ \}}
\textcolor{purple}{ "metric": \{}
\textcolor{purple}{   "objective_value": 0.500024080276,}
\textcolor{purple}{ \}}
\textcolor{purple}{\}}
\textcolor{purple}{"trial": \{}
\textcolor{purple}{ "parameter": \{}
\textcolor{purple}{   "minsplit": 8.0,}
\textcolor{purple}{   "minsplit.na": 0.0,}
\textcolor{purple}{   "minbucket": 32.0,}
\textcolor{purple}{   "cp": 0.30972302652187583,}
\textcolor{purple}{   "maxdepth": 4.0,}
\textcolor{purple}{   "maxdepth.na": 0.0,}
\textcolor{purple}{ \}}
\textcolor{purple}{ "metric": \{}
\textcolor{purple}{   "objective_value": 0.50002408028,}
\textcolor{purple}{ \}}
\textcolor{purple}{\}}
\textcolor{purple}{...}
    \end{Verbatim}
    \end{minipage}\\
    \bottomrule
    
    \end{tabular}
    }
\end{table}

\clearpage 

\section{Model and training details}
The open-sourced T5 model codebase we use can be found at \url{https://github.com/google-research/t5x}. 
\subsection{Conditional probability decomposition}
\label{appendix:conditional_probabilities}
From \cref{subsec:model}, the joint distribution of the optimization history $h$ conditioned on metadata $m$ can be written using the chain rule as

\begin{align}
    &P(\bar{\vh}|\bar{m}) 
    = P\left(\bar{x}_1^{(1)}, \bar{x}_1^{(2)}, \dots, \bar{x}_1^{(D)}, \star, \bar{y}_1, \text{"|"}, \dots,
          \bar{x}_T^{(1)}, \bar{x}_T^{(2)}, \dots, \bar{x}_T^{(D)}, \star, \bar{y}_T | \bar{m} \right) \nonumber\\
    &= \prod_{t=1}^T \left(\prod_{d=1}^D P\left(\bar{x}_t^{(d)}|\bar{m}, \bar{\vh}_{t-1}, \bar{\vx}_{t}^{(1:d-1)}\right)\right) 
                     P\left(\star|\bar{m}, \bar{\vh}_{t-1}, \bar{\vx}_{t}\right)
                     P\left(\bar{y}_t|\bar{m}, \bar{\vh}_{t-1}, \bar{\vx}_{t}\right)
                     P\left(\text{"|"}|\bar{m}, \bar{\vh}_t\right)
\end{align}

We note that this correctly formalizes the prediction of objects we are most interested in, which are parameter values $P\left(\bar{x}_t^{(d)}|\bar{m}, \bar{\vh}_{t-1}, \bar{\vx}_{t}^{(1:d-1)}\right)$ and function values $P\left(\bar{y}_t|\bar{m}, \bar{\vh}_{t-1}, \bar{\vx}_{t}\right)$.

\subsection{Training}
\label{sec:training_details}
During training, the encoder (denoted as $\mathbf{E}_{\theta}$) input sequence length is selected to be the maximum length of the tokenized metadata $\bar{m}$ from a dataset, ranging from 256 to 1024. The decoder (denoted as $\mathbf{D}_{\theta}$) input sequence is fixed at 1024, which means it can model up to $1024 // (D + 3)$ trials where $D$ is the number of parameters per trial. We use Adam optimizer with a rsqrt learning rate schedule and a mini-batch size of 256, and train each model up to 1M steps, with early stopping according to the validation loss. Each model is trained with a 4x4 TPU-v3 slice.

Thus the prediction for $\bar{h}^{(n)}$ is:
\begin{equation}
    P_{\theta}\left(\bar{h}^{(n)} \Big\vert m, \bar{\vh}^{(1:n-1)}\right) = \mathrm{SoftMax} \left[ \mathbf{D}_{\theta}(\mathbf{E}_{\theta}(\bar{m}), \bar{\vh}^{(1:n-1)}) \right] \label{eq:transformer}
\end{equation}

\subsection{Data augmentation}
We adopt the following three data augmentations to reduce overfitting to the offline datasets: 
\begin{enumerate}
    \item In order for the model to be invariant to parameter ordering, we apply random parameter permutations over metadata $\bar{m}$ and every suggestion $\bar{\vx}_{t}$.
    \item In order for the model to be robust to a different normalization range given a new function, we apply random scaling and shifting to the normalized function value $y_\mathrm{norm} = (y - y_\mathrm{min})/(y_\mathrm{max} - y_\mathrm{min})$ before quantization:
\begin{equation}
    y_\mathrm{norm}' = y_\mathrm{norm} * s + c,~s \sim \mathrm{Uniform}[0.3, 1],~c \sim \mathrm{Uniform}[0, 1-s]
\end{equation}
and thus $y_\mathrm{norm}' \in [c, c+s] \subseteq[0, 1]$ after transformation.
    \item Randomly drop textual and parameter value range information in metadata.
\end{enumerate}

\subsection{Inference}
\label{sec:inferencw_details}

At inference time, we choose the decoder input sequence length according to the maximum number of trials to run. E.g. to optimize a function with 18 parameters (highest possible dimension $D$ over our test functions) over 105 trials, we set the input sequence length to be at least $(18 + 3) * 105 = 2205$.

We compute the $(y_\mathrm{min}, y_\mathrm{max})$ range for function value normalization in the tokenization process with the current minimal and maximum observations. We set $c=0.2, s=0.6$ so that all normalized observations fall in the range of $y'_\mathrm{norm} \in [0.2, 0.8]$, and the model's $y$ value predicted distribution support, $[0, 1]$, is sufficiently large.

We also use a softmax temperature hyperparameter when predicting function values. We choose the temperature to maximize the log-likelihood of the validation split of each dataset seperately. On \vizierdata, the function prediction temperature is set as 1.1 and on \hpobdata it is 1.5. The policy prediction temperature is always set to be $1$.

\clearpage 

\section{Additional experimental results}

We provide additional experimental results in this section.

\subsection{Imitating HPO policies}
\label{sec:more_exp_imitate}

\begin{figure}[h]
    \centering
    \includegraphics[width=\textwidth]{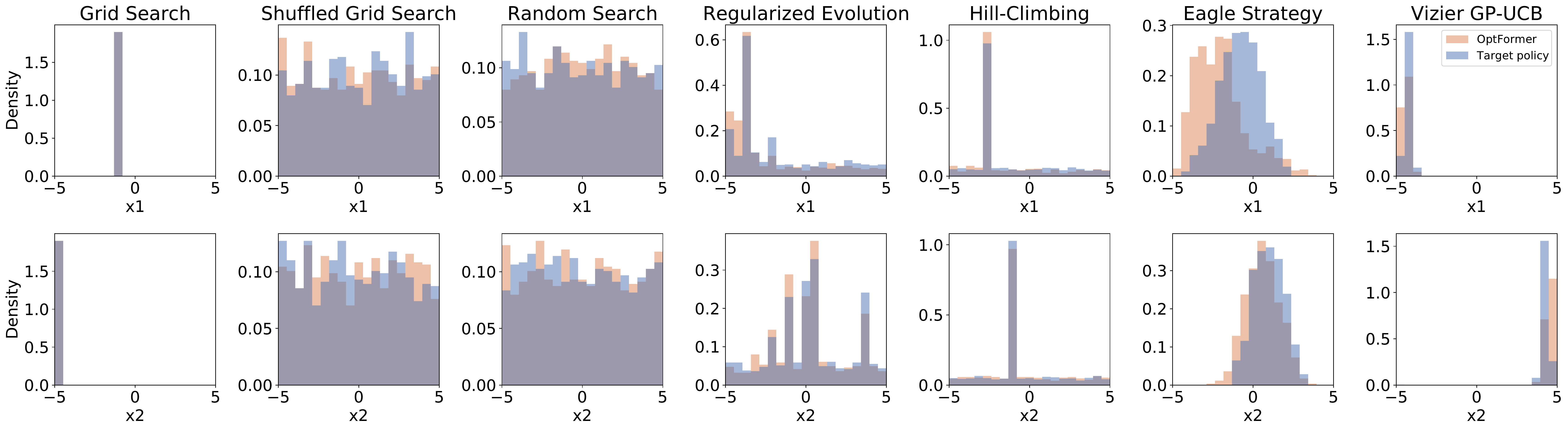}
    \caption{Policy distribution $p(x_{40}^{(d)} | m, \vh_{39}, \vx_{40}^{(1:d-1)})$ for $d=1, 2$ on a 2D GRIEWANK ROSENBROCK function.}
    \label{fig:x_pred_at_trial_40_2D_GRIEWANK}
\end{figure}

\begin{figure}[h]
    \centering
    \includegraphics[width=\textwidth]{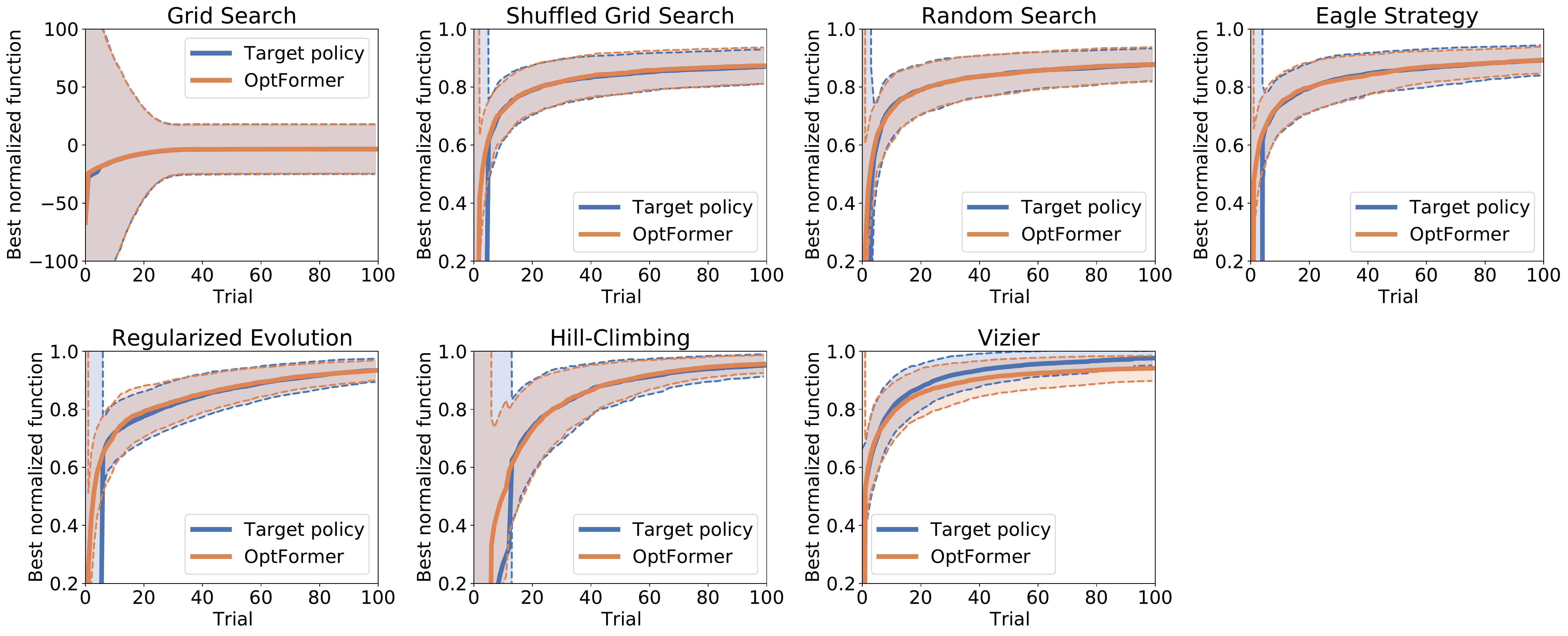}
    \caption{Best normalized function value with std, averaged over 5 test functions each with 100 runs.}
    \label{fig:bbob_opt_curve_per_alg_AVERAGED}
\end{figure}
\begin{figure}[h]
    \centering
    \includegraphics[width=\textwidth]{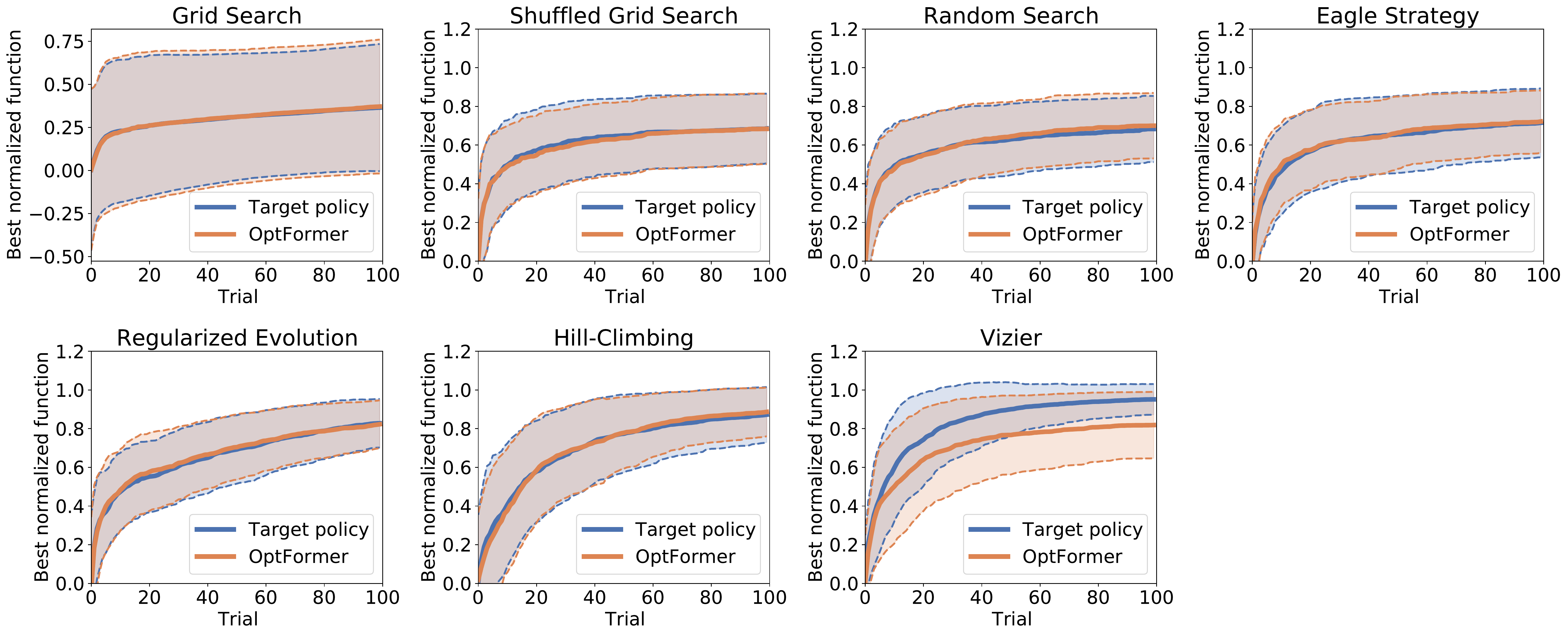}
    \caption{Best normalized function value of LINEAR SLOPE with std, averaged over 100 runs.}
    \label{fig:bbob_opt_curve_per_alg_LINEAR_SLOPE}
\end{figure}
\begin{figure}[h]
    \centering
    \includegraphics[width=\textwidth]{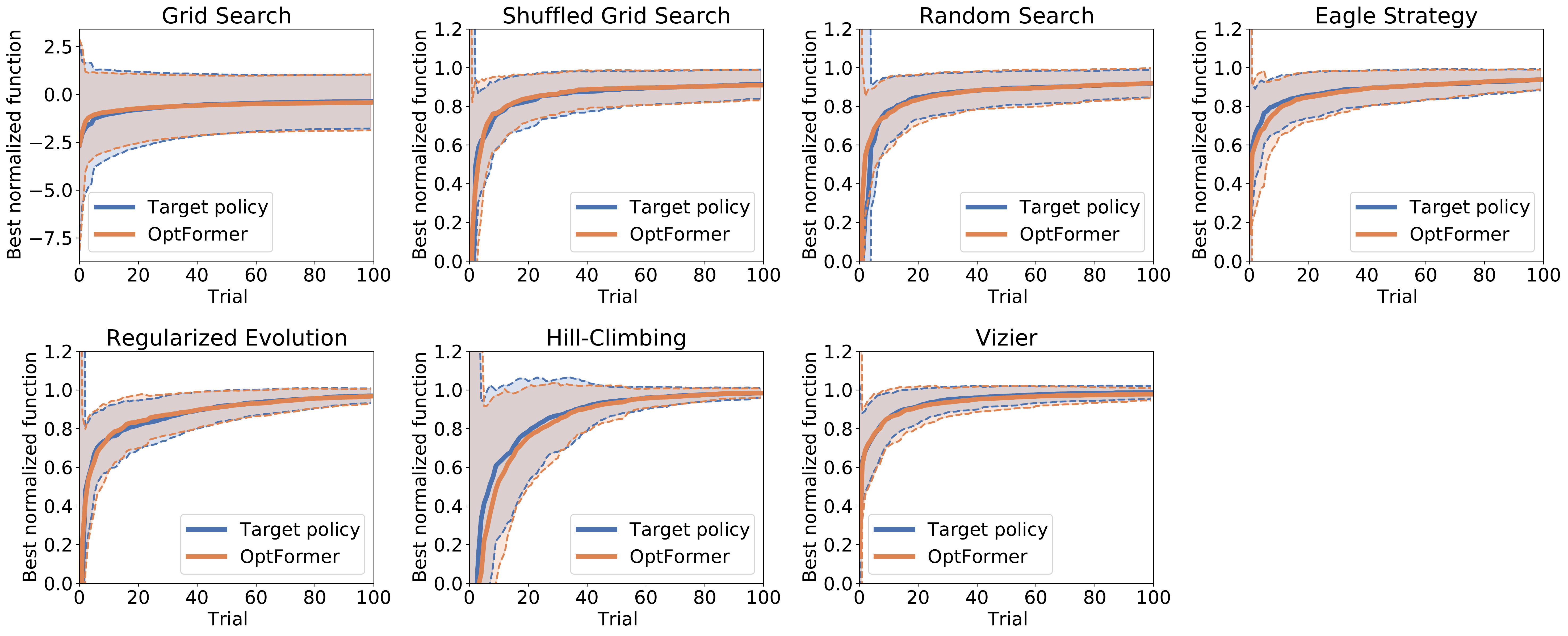}
    \caption{Best normalized function value of ROSENBROCK ROTATED with std, averaged over 100 runs.}
    \label{fig:bbob_opt_curve_per_alg_ROSENBROCK_ROTATED}
\end{figure}
\begin{figure}[h]
    \centering
    \includegraphics[width=\textwidth]{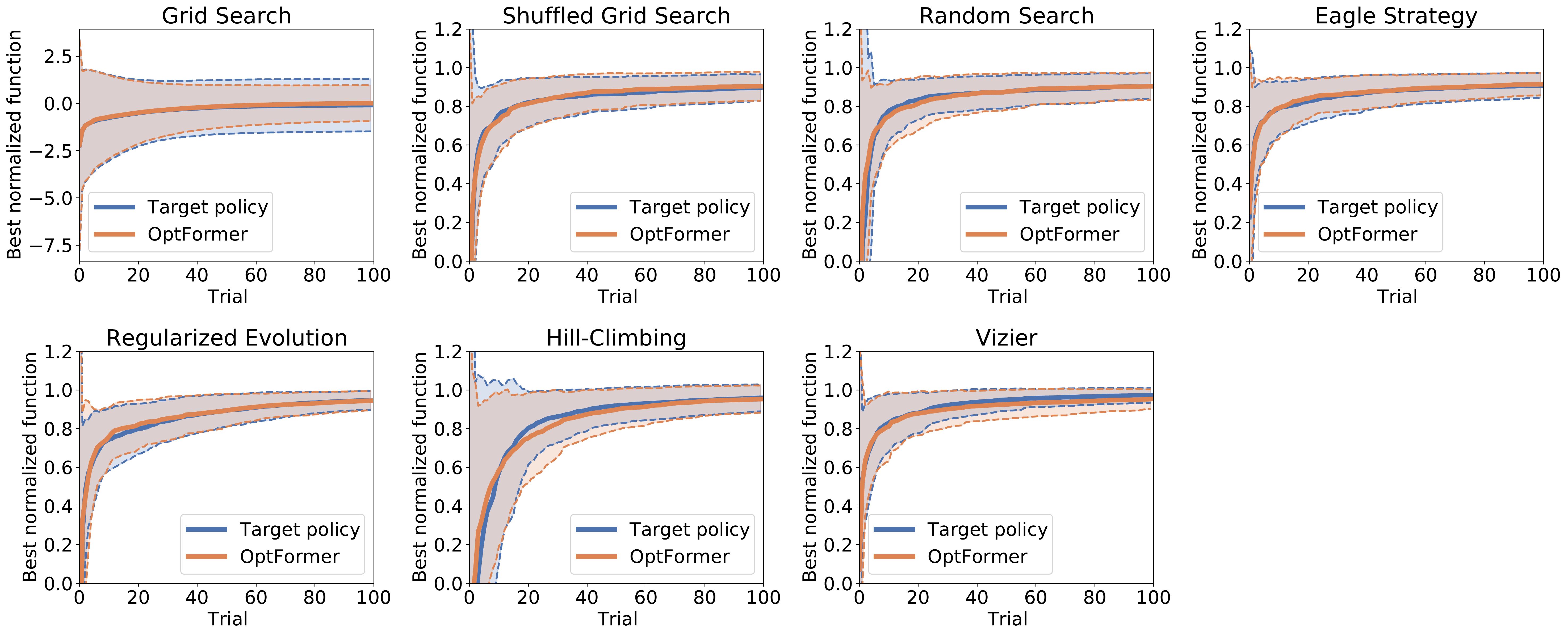}
    \caption{Best normalized function value of SUM OF POWERS with std, averaged over 100 runs.}
    \label{fig:bbob_opt_curve_per_alg_SUM_OF_POWERS}
\end{figure}
\begin{figure}[h]
    \centering
    \includegraphics[width=\textwidth]{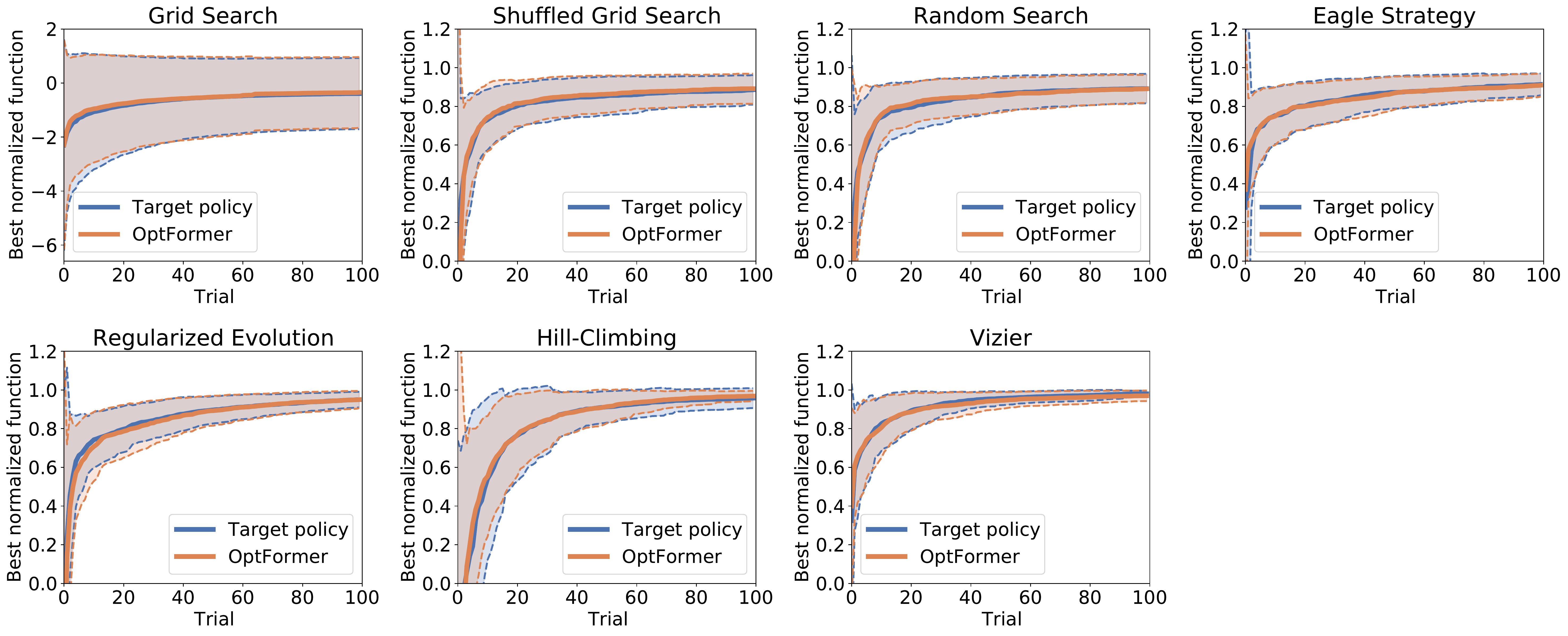}
    \caption{Best normalized function value of GRIEWANK ROSENBROCK with std, averaged over 100 runs.}
    \label{fig:bbob_opt_curve_per_alg_GRIEWANK_ROSENBROCK}
\end{figure}
\begin{figure}[h]
    \centering
    \includegraphics[width=\textwidth]{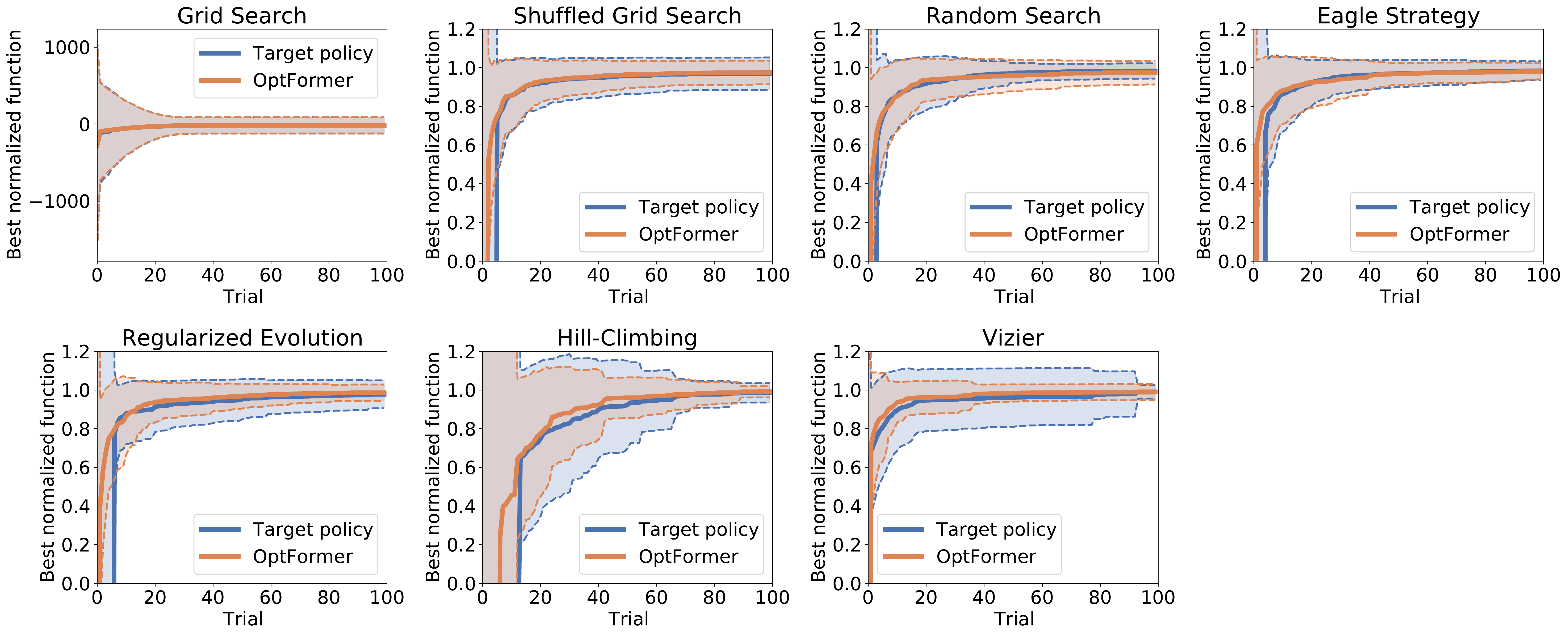}
    \caption{Best normalized function value for LUNACEK with std, averaged over 100 runs.}
    \label{fig:bbob_opt_curve_per_alg_LUNACEK}
    \vspace{128in}
\end{figure}
\clearpage

\subsection{Learning priors for hyperparameter response functions}
\label{sec:more_exp_fun_prior}
\begin{figure}[h]
    \centering
    \includegraphics[width=0.5\textwidth]{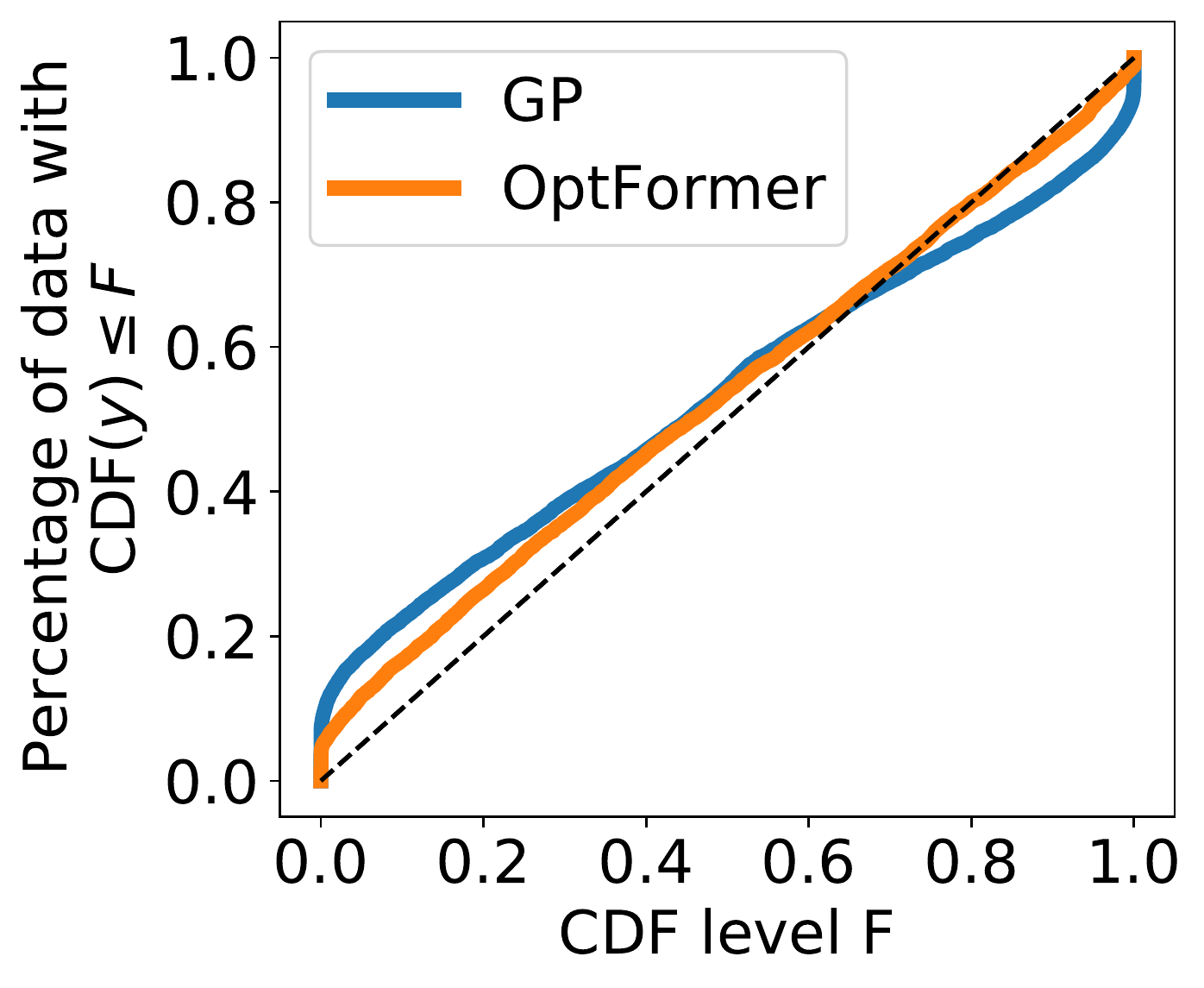}
    \captionof{figure}{Fitness of predicted $\mathrm{CDF(y)}$ on \hpobdata test set.}
    \label{fig:hpob_calibration}
\end{figure}

We apply the same goodness-of-fit analysis on function prediction from \cref{sec:fun_prior} to the test split of \hpobdata. The results are shown in \cref{fig:hpob_calibration}.

\vspace{0.2cm}
\vspace{0.2cm}

The ECE metric is defined for a classification model. To obtain a similar measurement for a continuous regression model, we convert the continuous regression problem into a multi-class classification problem by discretizing the range $[y_\mathrm{min}, y_\mathrm{max}]$ for each study into 100 equal intervals. Then, we follow the definition of ECE in \cite{naeini2015obtaining} and estimate the metric using 10 confidence bins.

\subsection{Augmenting a prior policy with function prediction}
\label{sec:more_exp_compare_policy}

\paragraph{Transfer learning results on HPO-B}
\cref{fig:y_curve_per_step} shows the best normalized function values observed so far at each trial. Though HyperBO uses a smaller NN for feature extraction, HyperBO has a flexible mean function, which captures important information that benefits BO in beginning trials. While we implemented a bare-bone FSBO, its performance is still better than ABLR in part thanks to FSBO's use of a squared exponential kernel instead of a dot-product one. Compared to a dot-product kernel with a finite feature space, a squared exponential kernel introduces infinite features.

\begin{figure}[h]
    \centering
    \includegraphics[width=\textwidth]{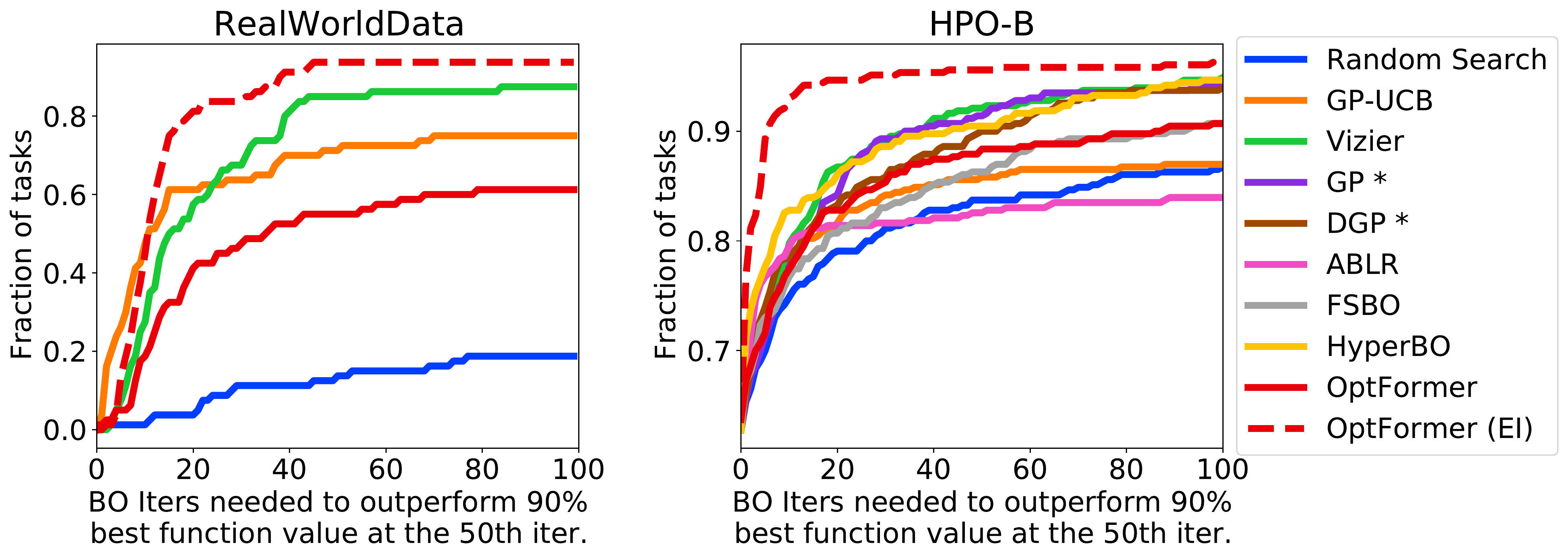}
    \caption{Performance profile on \vizierdata and \hpobdata test functions with success threshold: 90\% best function value at 50th iteration.}
    \label{fig:both_pp}
\end{figure}

\begin{figure}[h]
    \centering
    \includegraphics[width=\textwidth]{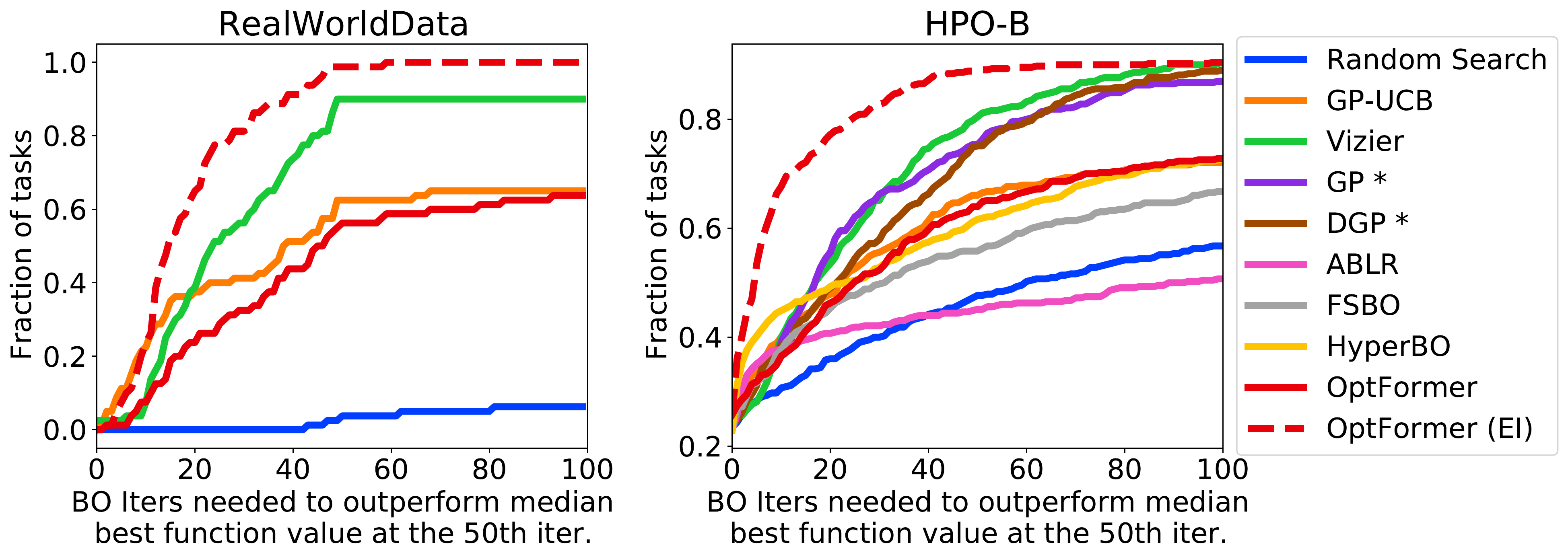}
    \caption{Performance profile on \vizierdata and \hpobdata test functions with success threshold: median best function value at 50th iteration.}
    \label{fig:both_pp_median}
\end{figure}

In \cref{fig:both_pp} and \cref{fig:both_pp_median}, we show the performance profiles of all compared methods over 2 different metrics: outperforming 90\% of the best function value obtained by all methods at the 50th iteration, and outperforming the median of the best function values obtained by each method at the 50th iteration.

Performance profiling is a performance evaluation tool to compare optimization methods, which is widely used in optimization~\cite{dolan2002benchmarking}. In our case, the y-axis is the fraction of tasks that each method succeeds in at different BO iterations (x-axis). The criteria of success depends on the problem itself, and we present performance profiles based on 2 different metrics: outperforming 90\% of the best function value obtained by all methods at the 50th iteration, and outperforming the median of the best function values obtained by each method at the 50th iteration.

Despite the relatively better performance of HyperBO, FSBO, and ABLR especially during earlier trials as shown by \cref{fig:y_curve_per_step}, these methods do not achieve a high percentage success rate on the 86 HPO-B test functions as reflected by \cref{fig:both_pp_median}. As pointed out by \citet{wang2021hyperbo}, ABLR, FSBO can be viewed as special cases of HyperBO with specific settings of kernel and mean functions. These methods have guarantees only if each function (corresponding to each task) is an i.i.d. sample from the same GP. However, for some search spaces in \hpobdata, there exist surrogate functions that return constant values. The constant surrogate function is unlikely to be an i.i.d. sample from the same GP as other surrogates in the same search space. This means ABLR, FSBO, and HyperBO can be sensitive to how the data is generated and outliers in the training data.

Summarizing the results in \cref{fig:y_curve_per_step}, \cref{fig:both_pp} and \cref{fig:both_pp_median}, HyperBO is able to achieve very good overall performance on a subset of all search spaces, which leads to a better averaged best normalized function values. It is likely that these search spaces have surrogate functions that meet the i.i.d function sample assumption from~\citet{wang2021hyperbo}. However, if we only look at the fraction of tasks each method surpasses a success metric, HyperBO may not be a method with superior performance that is comparable to the \model. This reveals another benefit of the \model: robustness to function outliers.


\paragraph{HPO-B plotting} We further compare the augmented policies from \cref{sec:compare_policy} to the provided baselines for \hpobdata in \citep{hpo-b}, using the same plotting format from \citep{hpo-b} for fair comparison.

\begin{figure}[h]
    \centering
    \begin{subfigure}[b]{0.6\textwidth}
        \centering
        \includegraphics[width=\textwidth]{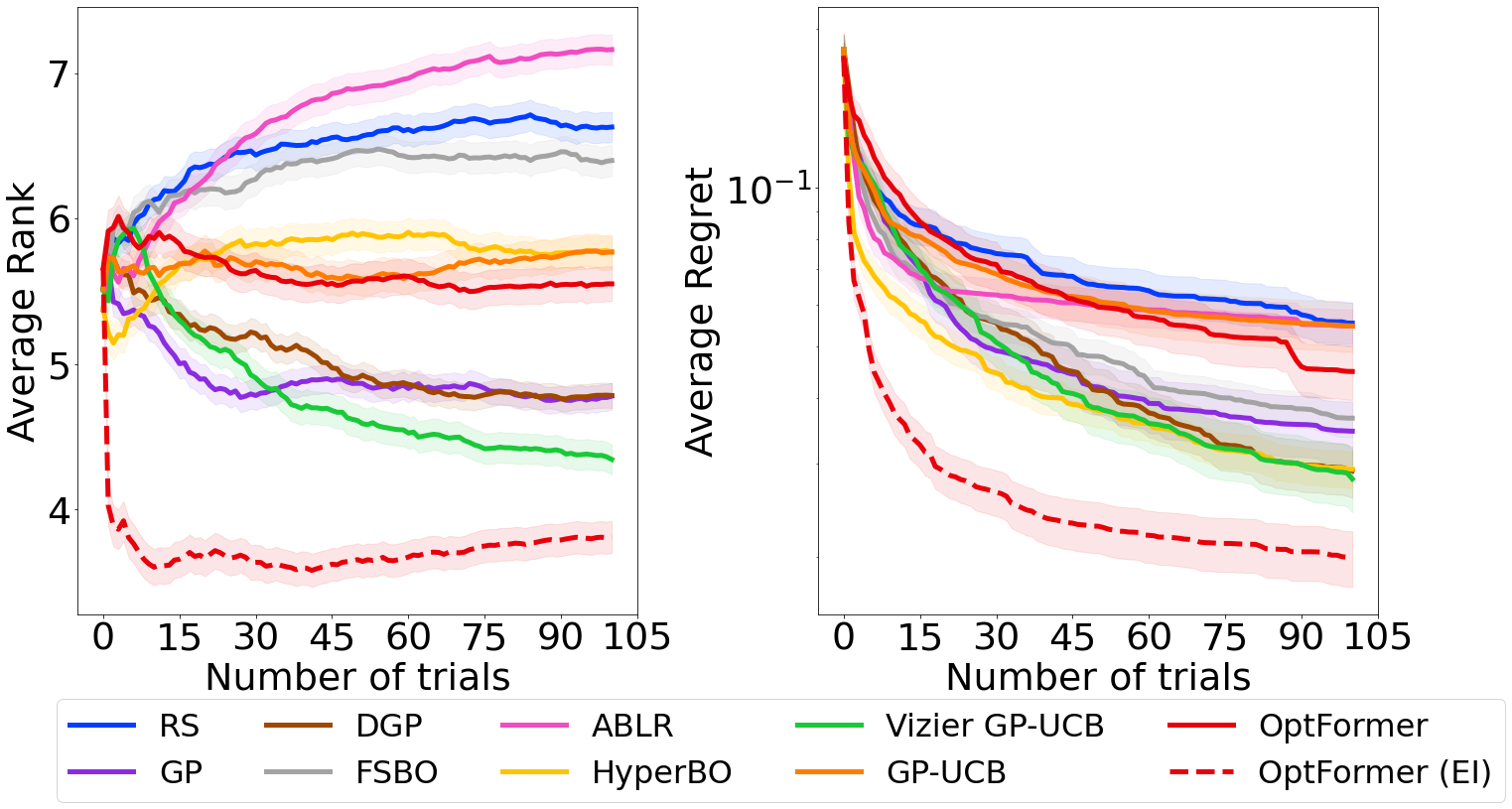}
    \end{subfigure}
    \hfill
    \begin{subfigure}[b]{0.38\textwidth}
        \centering
        \includegraphics[width=\textwidth]{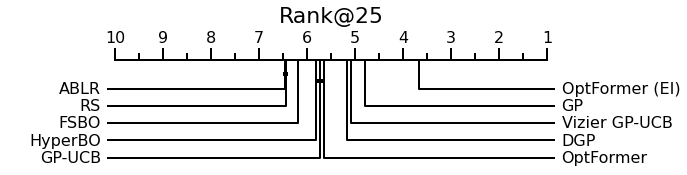}\\
        \includegraphics[width=\textwidth]{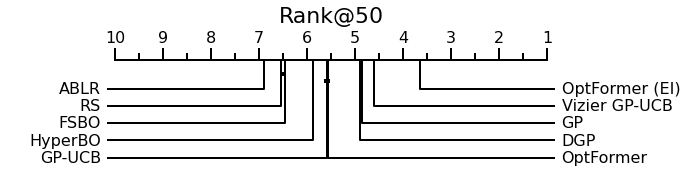}\\
        \includegraphics[width=\textwidth]{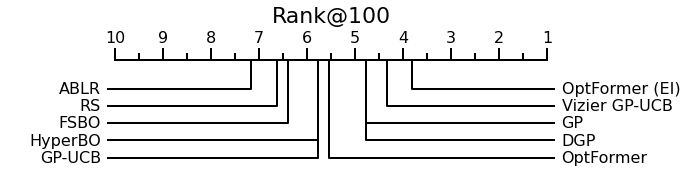}
    \end{subfigure}
    \caption{(Lower is better) Aggregated comparisons of normalized regret and mean ranks across all search spaces on the continuous search spaces of HPO-B-v3.}
    \label{fig:hpob_plotter}
\end{figure}

\subsection{Ablation on acquisition functions}
\label{sec:more_exp_ablation}

We provide additional ablations on acquisition function choices on both the \vizierdata and \hpobdata datasets.

\begin{figure}
    \centering
    \includegraphics[width=\textwidth]{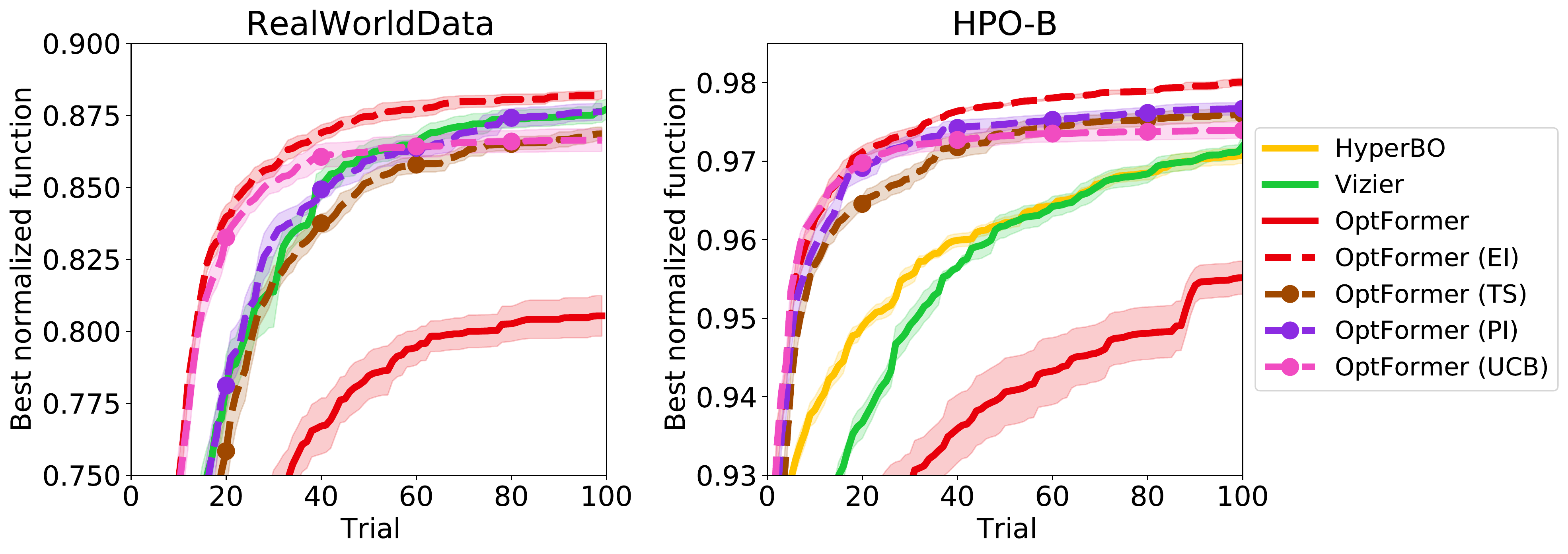}
    \caption{Ablation on the choice of acquisition functions. The plot shows the best normalized function values averaged over \hpobdata test functions. Ablation curves are shown with $\bigcirc$ markers.}
    \label{fig:ablation_acquisition_both_y}
\end{figure}

In \cref{fig:ablation_acquisition_both_y}, we compare the Expected Improvement (EI) used in the main body with Thompson Sampling (TS), Probability of Improvement (PI), and Upper Confidence Bound (UCB) with a confidence level of 0.9. We also include the best performing standalone baseline, Vizier, and transfer learning baseline, HyperBO, for reference.
We observe that the prior policy is improved by all the acquisition functions. Particularly, \model (EI) is the best among all acquisition functions and clearly outperforms all the baseline methods (HyperBO and Vizier) on both datasets across all trial steps. \model (UCB) finds good parameter settings as quickly as EI initially, but then becomes saturated early, suggesting a less exploratory behavoir than EI. The performance of PI and TS increases more slowly, but keeps improving compared to UCB.

\begin{figure}
    \centering
    \includegraphics[width=\textwidth]{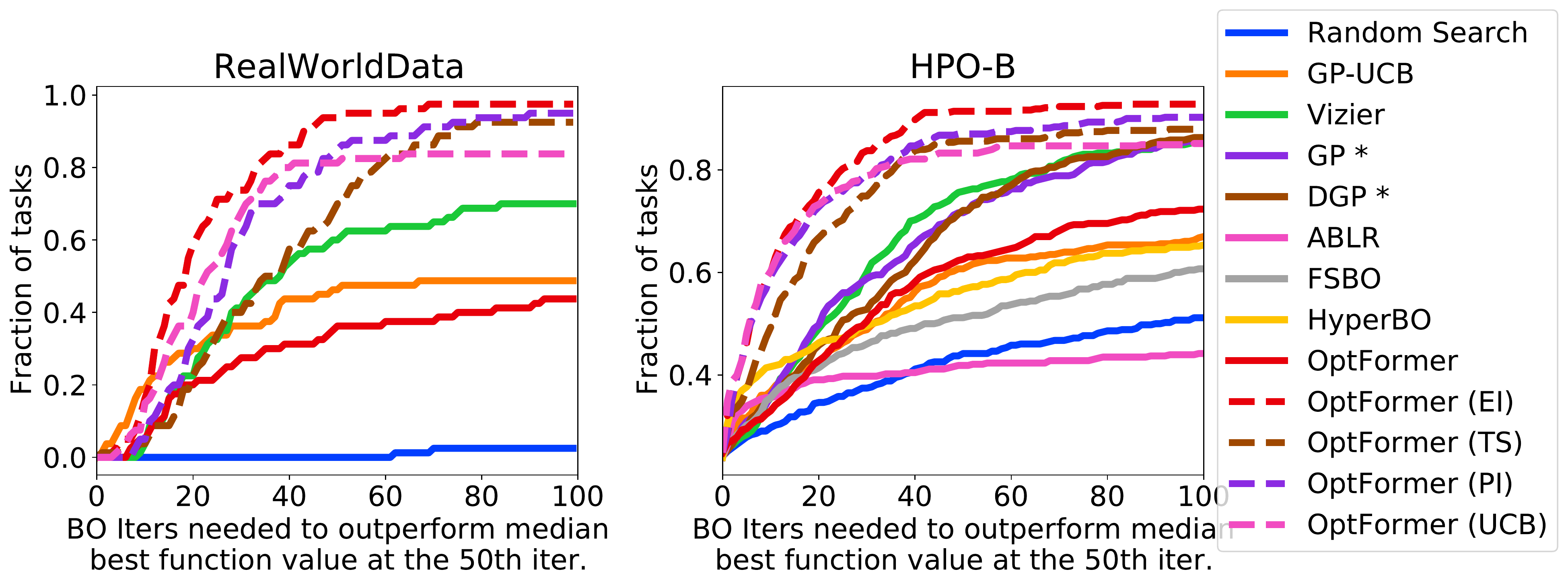}
    \caption{Ablation on the choice of acquisition functions. The plot shows the performance profile metric with success threshold: median best function value at 50th iteration.}
    \label{fig:ablation_acquisition_pp}
\end{figure}

To further bolster this hypothesis, we also compare using performance profiles.  As this metric depends on the set of methods being compared, we include all baselines from the main body. As we can see, \cref{fig:ablation_acquisition_pp} demonstrates that augmented \model policies, especially \model (EI), produce superior performance compared to other baselines.

\clearpage

\begin{figure}[h]
    \centering
    \includegraphics[width=\textwidth]{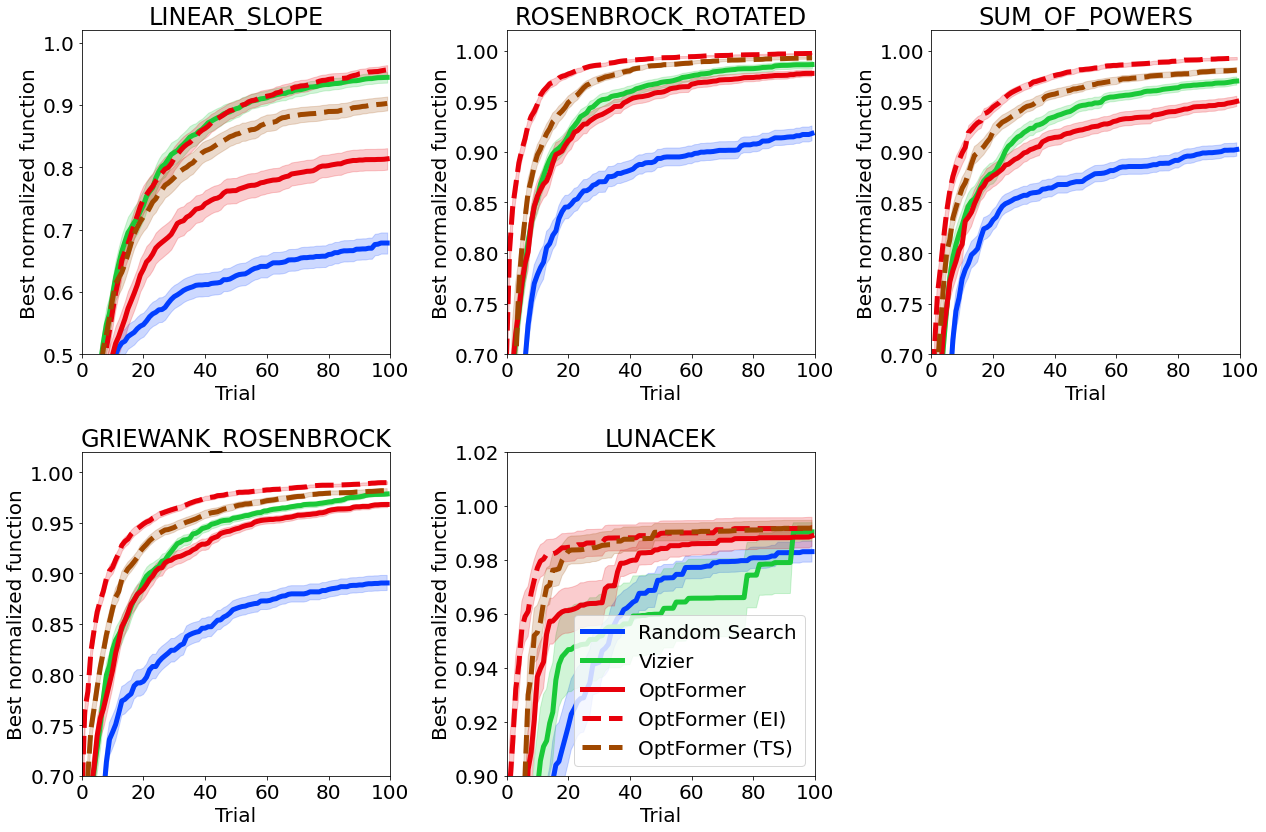}
    \captionof{figure}{Best normalized function value of a test function in \bbobdata averaged over 100 runs with std of the mean estimate.}
    \label{fig:bbob_opt_curve_per_fun_optformer_ei_ts}
\end{figure}

\edit{
\subsection{Out-of-Distribution functions}
\label{sec:out_of_distribution_comparisons}
}
\edit{
\cref{fig:bbob_opt_curve_per_fun_optformer_ei_ts} compares the optimization trajectories of the prior policy \model, augmented policies with EI (\model (EI)) and Thompson Sampling (\model(TS)), against Vizier and Random Search on 5 hold-out test function families from the \bbobdata benchmark. This assesses their performance on a few commonly used test functions for general black-box optimization. Both variants of the augmented policy obtain comparable or better performance than Vizier on most test functions except \model (TS) on the family of Linear Slope functions.
}

\begin{figure}[h]
    \centering
    \includegraphics[width=0.8\textwidth]{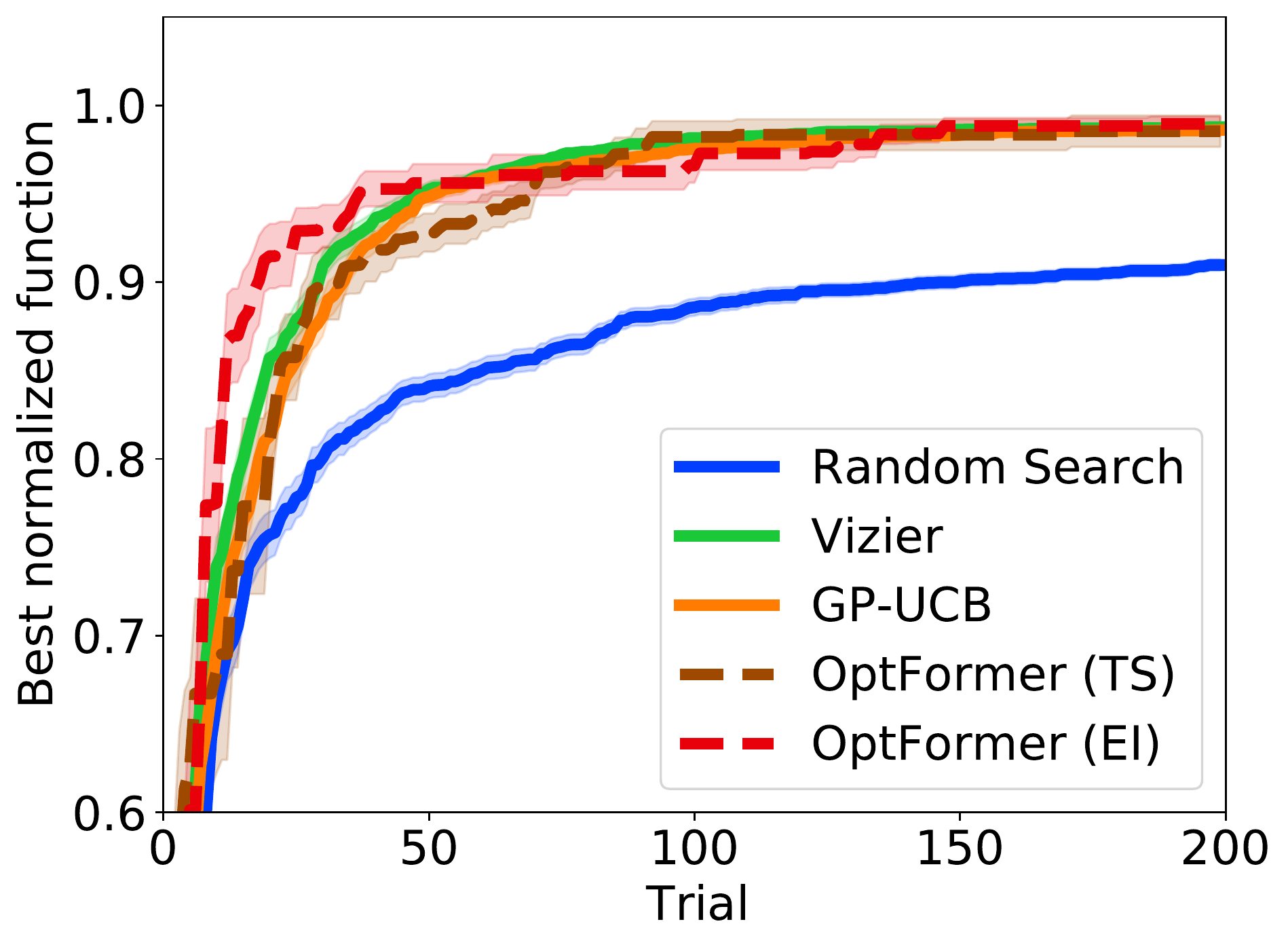}
    \captionof{figure}{Best normalized function value of NASBench averaged over 10 runs with std of the mean estimate.}
    \label{fig:nasbench_opt_curve}
\end{figure}

\edit{
In \cref{fig:nasbench_opt_curve} and \cref{fig:cifar10_init2_winit}, we further ablate the OptFormer on two machine learning tuning tasks: neural architecture search via NASBench-201 \citep{nasbench201} and tuning the learning rate schedule hyperparameters over a live CIFAR-10 training setup using a ResNet-50 from the init2winit benchmark \footnote{\url{https://github.com/google/init2winit}}. This assesses their performance on out-of-domain machine learning HPO tasks from the training datasets. Again, \model (EI) and \model (TS) perform comparably or even better than Vizier. This demonstrate their robust generalization performance over unseen tasks.
}

\edit{
\begin{figure}[h]
    \centering
    \includegraphics[width=0.8\textwidth]{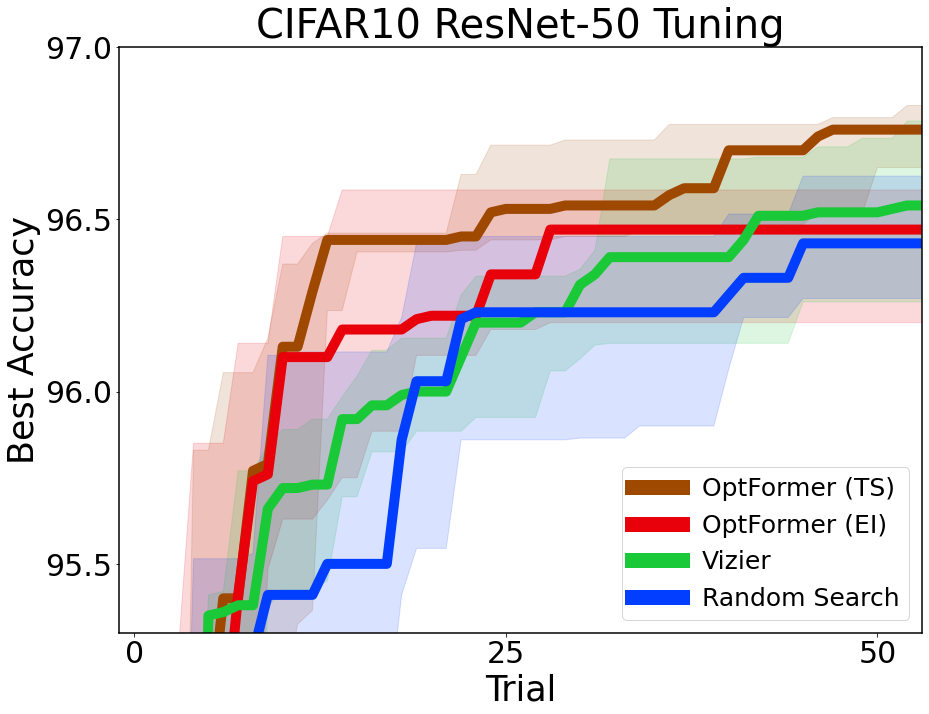}
    \captionof{figure}{Best CIFAR10 validation accuracy averaged over 10 runs with 25/50/75th percentiles shown.}
    \label{fig:cifar10_init2_winit}
\end{figure}
}

\end{document}